\documentclass[10pt,twocolumn,letterpaper]{article}

\usepackage{cvpr}              

\usepackage[dvipsnames]{xcolor}

\usepackage{graphicx}
\usepackage{subcaption}
\usepackage{float}
\usepackage{caption}
\usepackage{lscape}                                         

\usepackage{booktabs}                   
\usepackage{multirow}

\usepackage{bm}                          
\usepackage{times}
\usepackage{xcolor}
\usepackage{colortbl}
\usepackage{mathptmx}
\usepackage{mathtools}
\usepackage{amssymb,amsmath}   

\usepackage[export]{adjustbox}

\usepackage{tikz}
\usepackage{xcolor}
\colorlet{dark-blue}{blue!50!black}
\colorlet{dark-cyan}{cyan!75!black}
\colorlet{dark-purple}{purple!50!black}
\colorlet{dark-red}{red!75!black}
\colorlet{dark-green}{green!80!black}
\colorlet{dark-orange}{orange!50!black}
\colorlet{dark-gray}{black!75}
\colorlet{light-gray}{black!30}
\definecolor{nice-red}{HTML}{E41A1C}
\definecolor{nice-orange}{HTML}{FF7F00}
\definecolor{nice-yellow}{HTML}{FFC020}
\definecolor{nice-green}{HTML}{39b54a}
\definecolor{nice-blue}{HTML}{0071bc}
\definecolor{nice-purple}{HTML}{984EA3}

\definecolor{disco-d}{HTML}{E74C3C}
\definecolor{disco-i}{HTML}{9B59B6}
\definecolor{disco-s}{HTML}{3498DB}
\definecolor{disco-c}{HTML}{2ECC71}
\definecolor{disco-o}{HTML}{F1C40F}

\definecolor{darkgreen}{HTML}{3F7D31}
\definecolor{darkred}{HTML}{BA3132}

\definecolor{CBLightGreen}{HTML}{C5CFFF}

\definecolor{second}{rgb}{1, 0.85, 0.7}
\definecolor{best}{rgb}{1, 0.7, 0.7}
\definecolor{third}{rgb}{1,1, 0.8}

\colorlet{verylight-gray}{black!10}
\definecolor{LightCyan}{rgb}{0.66,0.85,0.76}
\newcommand{\tikzcircle}[2][red,fill=red]{\tikz[baseline=-0.7ex]\draw[#1,radius=#2] (0,0) circle ;}%
\definecolor{gold}{RGB}{255, 192, 0}
\definecolor{silver}{RGB}{215, 215, 215}
\definecolor{bronze}{RGB}{126, 66, 5}

\newcommand{\gold}{\tikzcircle[nice-red,fill=gold]{2pt}}

\newcommand{\heading}[1]{\noindent\textbf{#1}}

\makeatletter
\def\@fnsymbol#1{\ensuremath{\ifcase#1\or \dagger\or \ddagger\or
\mathsection\or \mathparagraph\or \|\or **\or \dagger\dagger
\or \ddagger\ddagger \else\@ctrerr\fi}}

\newcommand{\printfnsymbol}[1]{%
  \textsuperscript{\@fnsymbol{#1}}%
}
\makeatother

\definecolor{cvprblue}{rgb}{0.21,0.49,0.74}
\usepackage[pagebackref,breaklinks,colorlinks,citecolor=blue]{hyperref}

\def\paperID{5256} 
\def\confName{CVPR}
\def\confYear{2024}

\DeclareMathAlphabet{\altmathcal}{OMS}{cmsy}{m}{n}
\newcommand{\Loss}{\bm{\altmathcal{L}}}

\title{
HumanNeRF-SE: A Simple yet Effective Approach to Animate HumanNeRF with Diverse Poses
}

\author{Caoyuan Ma$^{1}$ \quad 
Yu-Lun Liu$^2$ \quad 
Zhixiang Wang$^{3,4}$ \quad 
Wu Liu$^5$ \quad 
Xinchen Liu$^6$ \quad 
Zheng Wang$^{1}$\thanks{ Corresponding Author}\\
\normalsize{$^1$National Engineering Research Center for Multimedia Software, Institute of Artificial Intelligence, School of} \\ 
\normalsize{Computer Science, Wuhan University  \quad ${^2}$National Yang Ming Chiao Tung University} \quad ${^3}$The University of Tokyo\\
\normalsize{ ${^4}$National Institute of Informatics \quad  ${^5}$School of Information Science and Technology,} \\
\normalsize{ University of Science and Technology of China}
\quad $^6$JD Explore Academy
}

\begin{document}

\twocolumn[{%
\renewcommand\twocolumn[1][]{#1}%
\vspace{-6mm}
\maketitle
\begin{center}
    \centering
    \vspace{-8mm}
    \includegraphics[width=0.95\textwidth]{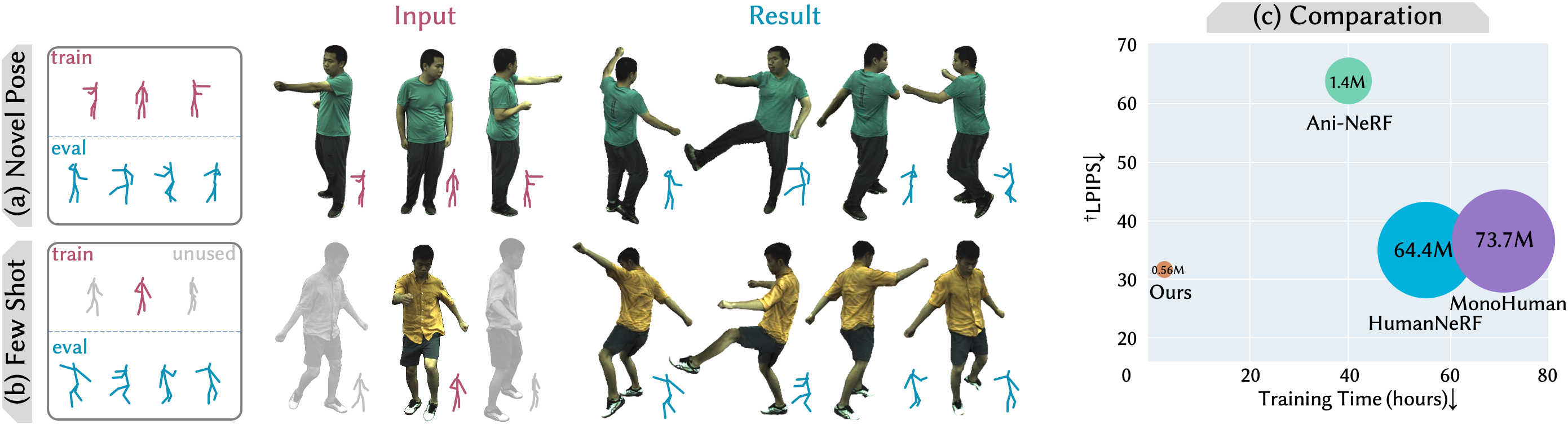}
    \captionof{figure}{
    \textbf{Overview.} 
    HumanNeRF-SE efficiently synthesizes images of performers in \emph{diverse} poses, blending \emph{simplicity} with \emph{effectiveness}.
    It outperforms previous methods by creating a wider range of new poses (a), maintains generalization without overfitting with limited input frames (b), and requires fewer than 1\% of learnable parameters, reducing training time by 95\% while delivering superior results in the few-shot scenario (c).
    $^\dagger$LPIPS = 1,000$\times$LPIPS. 
    Project page: \url{https://miles629.github.io/humanNeRF-se.github.io/}
    }
    
    \label{fig:teaser}
\end{center}%
}]

\begin{abstract}
We present HumanNeRF-SE, a simple yet effective method that synthesizes diverse novel pose images with simple input.
Previous HumanNeRF works require a large number of optimizable parameters to fit the human images. 
Instead, we reload these approaches by combining explicit and implicit human representations to design both generalized rigid deformation and specific non-rigid deformation.
Our key insight is that explicit shape can reduce the sampling points used to fit implicit representation, and frozen blending weights from SMPL constructing a generalized rigid deformation can effectively avoid overfitting and improve pose generalization performance.
Our architecture involving both explicit and implicit representation is simple yet effective. 
Experiments demonstrate our model can synthesize images under arbitrary poses with few-shot input and increase the speed of synthesizing images by 15 times through a reduction in computational complexity without using any existing acceleration modules. 
Compared to the state-of-the-art HumanNeRF studies, HumanNeRF-SE achieves better performance with fewer learnable parameters and less training time.

\end{abstract}    
\section{Introduction}
\label{sec:intro}

\begin{figure*}
    \centering
    \includegraphics[width=\linewidth]{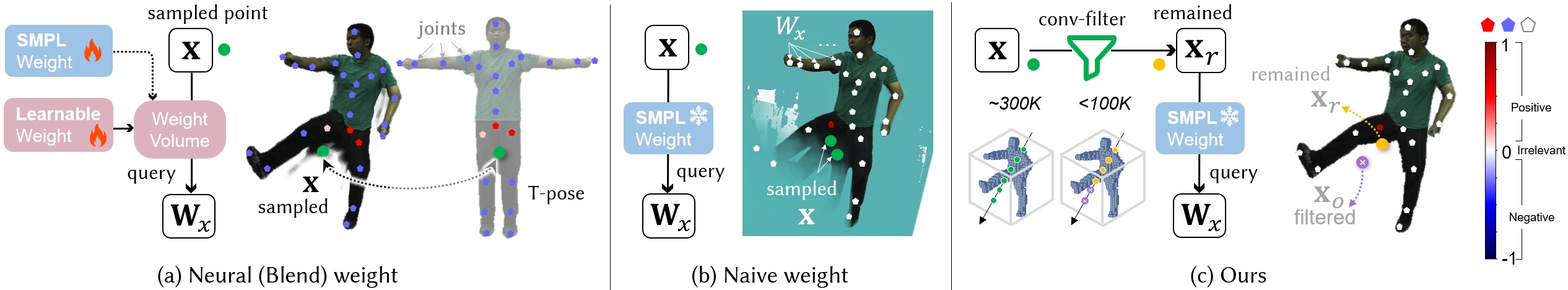}
    \vspace{-4mm}
    \caption{
    \textbf{Different weights for deformation.} 
    (a) Prior methods~\cite{peng2021animatable,weng2022humanNeRF,yu2023monohuman,liu2021neural} learn a weight volume for deformation through neural networks or fine-tune blending weights obtained from fitting SMPL to the input frame.
    The weight volume optimized along with NeRF parameters per human image is prone to over-fitting.
    When synthesizing novel pose images, the over-fitted weights will deform points onto the canonical space \emph{incorrectly} and lead to artifacts.
    (b) Our idea is to use SMPL's blending weights directly because these weights are pre-trained on numerous human images to avoid overfitting. However, simply utilizing the nearest SMPL vertex's blending weights for deformation fills the sampling space with incorrect colors as the training phase deforms irrelevant sampling points onto the human body.
    (c) We propose to filter irrelevant points according to the human body information of SMPL. This way, we can avoid over-fitting and reduce the number of sampling points.
    }
    \label{fig:motivation}
\end{figure*}



Neural Radiance Field (NeRF)~\cite{mildenhall2021NeRF,wang2021ibrnet,zhang2020NeRF++,wu2022dof,wang2022NeRF} represents the scene as an implicit field and utilizes volumetric renderer to synthesize the scene has demonstrated remarkable advancements in reconstruction and \emph{novel-view} syntheses of \emph{static} scenes.
However, they typically do not account for object deformation and perform poorly on \emph{dynamic} humans due to the complex deformation caused by motions.
Deformable NeRFs endow implicit fields with the capability to express dynamic objects~\cite{park2021NeRFies,park2021hyperNeRF,pumarola2021d,tretschk2021non,xian2021space} or even humans~\cite{jiang2022neuman,weng2022humanNeRF,yu2023monohuman,jiang2023instantavatar,geng2023learning}. 
Although these methods~\cite{jiang2022neuman,weng2022humanNeRF,yu2023monohuman,jiang2023instantavatar,geng2023learning} could learn high-quality human representations and synthesize images from arbitrary viewpoints, they cannot synthesize images with \emph{novel poses} that are significantly {different} from that of the training videos. 

We aim to automatically render photo-realistic human images with arbitrary \emph{viewpoints} and \emph{poses} from monocular videos. Although there are some studies~\cite{kwon2021neural,peng2021animatable,peng2021neural,liu2021neural,zhao2022humanNeRF,chen2023uv} have attempted to learn animatable human representations by introducing neural blend weights or using UV-referenced coordinate systems,
their requirement for multi-camera data limits their practical applicability. The problem becomes practical with monocular inputs but highly \emph{ill-posed} due to the limited and patchy observations. 

Existing monocular-based methods~\cite{jiang2022neuman,weng2022humanNeRF,yu2023monohuman,jiang2023instantavatar,geng2023learning} usually decompose the human implicit field into rigid and non-rigid components, reducing the ill-posedness in joint optimization.
These two components deform sampling points from the observation space to the canonical one.
The non-rigid field is learned by a neural network conditioned on the human pose or frame index. On the contrary, the rigid field uses an explicit model---Linear Blending Skinning (LBS)---given blending weights learned from scratch or fine-tuning SMPL's~\cite{loper2015smpl} weights. Since the data for training NeRF is limited and the number of optimizable parameters is large, the blending weights could overfit the input data and yield unsatisfactory results, especially when the input poses become very restricted (Figure~\ref{fig:motivation}a).



In this paper, we present HumanNeRF-SE, which synthesizes novel pose images with \emph{tens} of simple inputs and \emph{a few} learnable parameters in \emph{hours}. 
Our approach distinguishes itself from previous methods like HumanNeRF~\cite{weng2022humanNeRF} by effectively leveraging prior knowledge provided by SMPL.
On the one hand, we use the blending weights from pre-trained SMPL \emph{without} any change for rigid deformation. This is because SMPL trained on numerous human data is generalizable to diverse humans and poses.
On the other hand, we employ the SMPL's vertices for sampling points.
The motivation is that we found simply using the blending weights is not enough (Figure~\ref{fig:motivation}b) since there are a lot of irrelevant human points in the volume. These points could be deformed to the human body to be reconstructed incorrectly.
We propose the Conv-Filter guided by SMPL's vertices to reduce the irrelevant points (Figure~\ref{fig:motivation}c). Our method not only avoids overfitting but also greatly reduces the required sampling points from $\sim$300K to less than 100K. Besides sampling, we also use spatial-aware features extracted from SMPL's vertices to condition non-rigid deformation.



Specifically, we first voxelize SMPL's vertices by employing a sparse convolution to diffuse the vertices across the voxel volume. 
Second, the spatial-aware feature and occupancy of the sampling point can be easily queried in this volume.  
Throughout this procedure, a significant number of points unrelated to the body are filtered out, leaving behind only those points likely to be related to the human body. 
Third, these points can be deformed to a unified canonical pose using the rigid deformation to get the coarse coordinates.
Fourth, we refine the coarse coordinates by performing a non-rigid mapping conditioned on the point-level spatial-aware feature obtained in the convoluted voxel volume.
Finally, we get the colors and densities of the sampling points through a neural network and render the image through a differentiable renderer.

Overall, we propose a \emph{simple} yet \emph{effective} HumanNeRF method that synthesizes images of varying poses \emph{efficiently}. 
Our approach utilizes explicit SMPL prior knowledge to design a generalized rigid deformation and a specific non-rigid deformation to map points from observation space to canonical space. 
Our experiments show that our method can generate novel poses with significant differences from the training poses, even when the input is limited to a few shots.
We also further demonstrate the superiority of our method on our captured in-the-wild data, where the input video involves a simple rotation of the user.

To sum up, we make the following contributions:
\begin{itemize}
    \item We design modules to leverage the prior knowledge from SMPL for deformations and point sampling.  This dramatically reduces computational complexity and avoids overfitting.
    \item Our architecture is simple yet effective. Compared to methods with similar performance, HumanNeRF-SE uses less than 1\% learnable parameters, 1/20 training time, and increases rendering speed by 15 times without using any existing acceleration modules.
\end{itemize}

\section{Related Work}

\paragraph{3D Performance Capture.} Recently, deep neural networks are commonly used to learn scene or human representation from images, with a range of methods now available including voxels~\cite{sitzmann2019deepvoxels,lombardi2019neural}, point clouds~\cite{wu2020multi,aliev2020neural,habermann2020deepcap,xu2018monoperfcap}, textured meshes~\cite{thies2019deferred,jiang2022selfrecon,liu2019neural,liao2020towards,zhou2021dc,zheng2023avatarrex,saito2019pifu,saito2020pifuhd,xiu2022icon}, multi-plane images~\cite{zhou2018stereo,flynn2019deepview}, and implicit functions~\cite{sitzmann2019scene,liu2020dist,pumarola2021d,niemeyer2020differentiable,mildenhall2021NeRF,liu2020neural}. 
Most of these methods aim to optimize a detailed 3D geometry, and then synthesize results into images or videos for application, which limited their usage. Our method can directly synthesize human body images.

\paragraph{Neural Rendering.} To synthesize novel view images of a static object without recovering detailed 3D geometry, previous neural rendering method~\cite{barron2021mip,martin2018lookingood,martin2021NeRF,wang2021ibrnet,zhang2020NeRF++,hedman2021baking,niemeyer2021giraffe,srinivasan2021nerv,tancik2020fourier,zhang2021NeRFactor,wu2022dof,wang2022NeRF,li2024unleashing} represent amazing image synthesis result, but these methods typically assume multi-camera input and usually don't take object's deformation into consider. 
The deformable methods~\cite{park2021NeRFies,park2021hyperNeRF,pumarola2021d,tretschk2021non,gao2021dynamic,li2021neural,xian2021space} endow the implicit field with the ability to express dynamic objects but don't perform well in the human because of the complexity of human deformation. Our method employs a simple and effective architecture for human body deformation.
The ability to synthesize the movements of specific human bodies in different poses has a wide range of applications. Therefore, it is very meaningful to extend methods to adapt to dynamic human bodies.

\paragraph{Multi-camera HumanNeRF.} 
There has been some research~\cite{cheng2022generalizable,kwon2021neural,peng2021animatable,peng2021neural,liu2021neural,zhao2022humanNeRF,jayasundara2023flexNeRF,li2023posevocab,gao2022mpsNeRF,chen2022geometry,chen2023uv} on learning dynamic human representation through NeRF from multi-camera images.
NeuralBody~\cite{peng2021neural} uses structured pose features generated from SMPL~\cite{loper2015smpl} vertices to anchor sampling points in any poses from sparse multi-camera videos, which inspired us to use spatial information of vertices. 
\cite{peng2021animatable,liu2021neural,zhao2022humanNeRF,jayasundara2023flexNeRF,li2023posevocab,gao2022mpsNeRF} transform sampling points of dynamic human to a canonical space for NeRF training. 
Because the information in monocular data is much more limited than in multi-camera data,
some of them have the ability to train with only monocular video, but they are not designed for monocular scenes and usually don't perform well in monocular data. 
Our method only requires few-shot input and also performs well in monocular data.

\paragraph{Monocular HumanNeRF.}
Since obtaining monocular videos is much easier than obtaining synchronized multi-view videos, it is significant to extend the capabilities of HumanNeRF to monocular videos. Inspired by~\cite{park2021NeRFies,park2021hyperNeRF,pumarola2021d}, which maps rays in dynamic scenes to canonical space, \cite{jiang2022neuman,weng2022humanNeRF,yu2023monohuman,jiang2023instantavatar,geng2023learning} introduce priors to regularize the deformation. 
\cite{peng2023intrinsicngp,jiang2023instantavatar,geng2023learning} greatly improve the speed of training and rendering by using more efficient spatial encoding methods~\cite{muller2022instant,chan2022efficient}, while these encoding methods don't perform well in terms of acceleration in our experiments.
\cite{su2021aNeRF} learn dynamic human bodies by modeling the relationship between sampling points and joints. 
\cite{jiang2022neuman,peng2021animatable} learns the blending field by extending the deformation weight of the closest correspondence from the SMPL mesh, which leads to a significant increase in computational cost.
HumanNeRF~\cite{weng2022humanNeRF} demonstrates amazing novel view results by decoupling the motion field, which uses a significant amount of neural networks to fit various modules and often takes a very long time to train.
Monohuman~\cite{yu2023monohuman} further improves the pose generalization performance of HumanNeRF by adding a reference image module and consist loss.
Our experiments aimed to explore the fundamental reasons why HumanNeRF performs poorly in pose generalization, and to achieve better results on more challenging data by rebuilding a simple yet efficient architecture with SMPL vertices to combine explicit and implicit human representation.

\begin{figure*}
    \centering
    \includegraphics[width=0.9\textwidth]{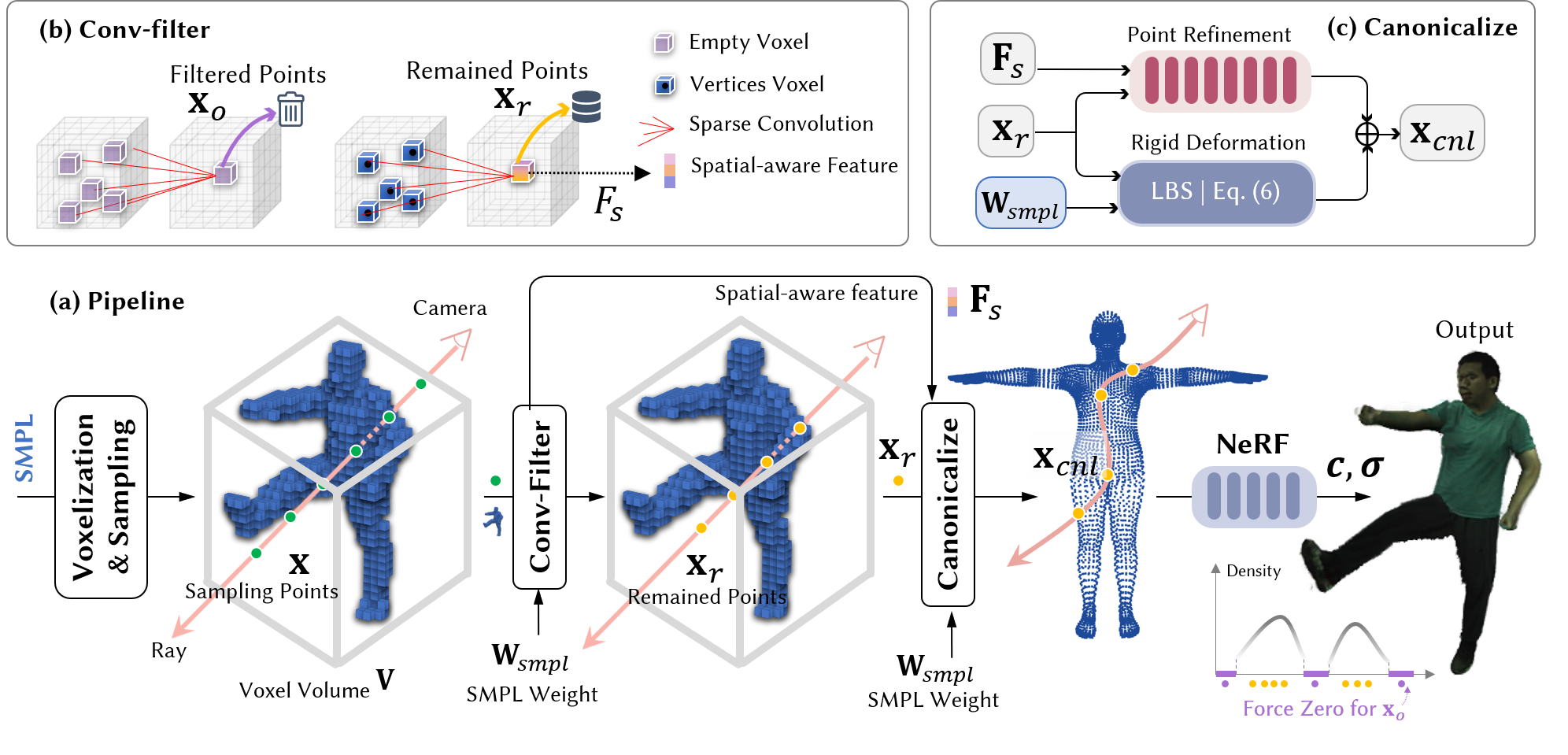}
    \vspace{-6mm}
    \caption{\textbf{Framework of HumanNeRF-SE.} 
    (a) We first voxelize the observation space as a voxel volume $\mathbf{V}$. For a voxel containing vertices, the value will be the number of vertices (as one occupancy channel) and the corresponding SMPL weight.
    (b) We performed channel-by-channel convolution on the volume. All sampling points are queried in the convolutional volume to get their spatial-aware features. Those points with zero occupancy will be filtered out. 
    (c) We query the nearest weight of the remained points in the volume, which is used for rigid deformation. Spatial-aware features are utilized in the neural network to correct the rigid results and obtain the final point coordinates in the canonical space.
    The sampling points in the canonical space obtain their colors and densities through the NeRF network. The densities of filtered points are forced to be zero.
    }
    \vspace{-3mm}
    \label{fig:method}
\end{figure*}

\section{HumanNeRF-SE}

We propose a simple yet effective approach for learning the implicit representation of human bodies from limited inputs and being capable of synthesizing diverse novel poses. 
Compared to other similar methods, each module of our approach is designed to map the explicit and implicit human representation better and generalize to arbitrary novel poses without overfitting.
This results in fewer data demands and computational load and improves pose generalization capabilities compared to other approaches. 
The pipeline is illustrated in Figure~\ref{fig:method}.

The key to learning the deformable NeRF representation of human bodies lies in canonicalizing the sampling points within the dynamic observation space. 
Prior methods predominantly depend on neural network fitting or supplementary texture information to precisely anchor the sampling points relative to the human body. 
These methods also introduce frame-level features to augment the multi-view results, albeit at the expense of substantially diminishing the pose generalization capabilities of the model. 

Different from these methods, our method effectively leverages the explicit vertices $\mathbf{v}$ to canonicalize the sampling points $\mathbf{x}$ within the dynamic human observation space. 
This process involves constructing voxel volume $\mathbf{V}$ (Sec.~\ref{sec:Voxelization}), convoluting voxel volume channel-by-channel to obtain spatially-aware features $F_s$ and filtering out useless sampling points $\mathbf{x}_o$ and get human-related points $\mathbf{x}_r$ (Sec.~\ref{sec:Conv-Filter}), mapping points to canonical space generally $D$ and specifically $P$ (Sec.~\ref{sec:Canonicalization}). 
It is important to emphasize that we exclusively utilize point-level features throughout the entire process, abstaining from the use of any frame-level features, thereby ensuring the pose generalization capabilities of our model. 

In summary, our method can be represented as:
\begin{equation}
    \mathbf{c},\sigma = M_{\sigma}(P(\mathbf{F}_{s},\mathbf{x}_r)+D(\mathbf{x}_r, K(\mathbf{x}_r,\mathbf{V}),J))
\label{eq:total}
\end{equation}
where $J$ represents the pose and $M_{\sigma}$ is the NeRF network which is similar to baselines for better comparison. We use $K$ to query the nearest weights of points $\mathbf{x}$ in the volume $\mathbf{V}$.
Our rigid deformation $D$ is a general mapping process, and it is not influenced by the training individual.

\subsection{Voxelization}
\label{sec:Voxelization}
 
In order to more efficiently handle the relationship between the sampling points and the SMPL model, we first voxelized the SMPL space. However, since the vertices of the SMPL model within the boundary of the human body are still relatively sparse, and we used the Sparse Convolution $Sp$~\cite{spconv2022} to construct our voxel volume~$\mathbf{V}$.

We processed data similarly to NeuralBody~\cite{peng2021neural} in this part. For a given set of SMPL vertices $\mathbf{v}$, we first calculate the maximum and minimum values on the coordinate axes to get the bounding box, scale the bounding box to the set voxel size $\mathbf{vs}$, and find the least common multiple axes and 32, in preparation for subsequent sparse convolution. 
For each SMPL vertex, we also scale it down according to voxel size after subtracting the minimum value.
\begin{equation}
    V = Sp(\mathbf{v}, \mathbf{vs}, \mathbf{W})
\end{equation}


In the generated voxel volume, each voxel that contains vertices holds two values: the corresponding LBS weight $w_j$ in SMPL weight $\mathbf{W}_{smpl}$ and the count of contained vertices $n_{v}$ as an occupancy indicator. For voxels without any contained vertices, all channels are assigned a value of zero. The value of a certain voxel is:
\begin{equation}
\left \{  
\begin{array}{ll}
(n_{v},w_1,w_2...,w_{24}), & \text{ if contain vertices}\\
(0,0,0...,0), &  \text{if empty voxel}
\end{array}
\right.  
\label{eq:voxelvalue}
\end{equation}


\subsection{Conv-Filter}
\label{sec:Conv-Filter}

We innovatively use spatial convolution to filter the sampling points and extract features simultaneously. 
A convolution kernel is initialized to one for convolving the value of the voxel volume. 
To preserve high-frequency information, we use channel-by-channel convolution.
\begin{equation}
    F_{s_i} = \sum_{m=h'}^{h} \sum_{n=w'}^{w} \sum_{t=d'}^{d} \boldsymbol{\nu}_{m,n,t,i} \cdot \vartheta_{m,n,t,i}
\end{equation}
where $h' = h-k$, $w'=w-k$ and $d'=d-k$. The $i$-th channel of $F_s$ results from convolving $i$-th channel of voxel's value $\boldsymbol{\nu}$ and kernel's weight $\vartheta$.
If the occupancy of the convolution result is zero, it means that the current convolution center coordinate is not related to the human body (Eq.~\ref{eq:voxelvalue}). We force these points not to participate in subsequent calculations and set their density value~$d$ to zero.
\begin{equation}
\left \{  
\begin{array}{lll}
\mathbf{x}_o \text{ filter out},\, d(\mathbf{x}_o) =0 & \text{if~} \mathbf{F}_{s}(x) = 0\\
\mathbf{x}_r \text{ remain}  &  \text{if~} \mathbf{F}_{s}(x) > 0 
\end{array}
\right.  
\end{equation}

\subsection{Point Canonicalization}
\label{sec:Canonicalization}

We compute the rotation $R_j$ and translation $T_j$ of the current pose relative to the joints in the T-pose of the human body in a similar way of HumanNeRF~\cite{weng2022humanNeRF}.
Inspired by NeuralBody~\cite{peng2021neural}'s use of convolution to diffuse vertex occupancy, we use a simpler method to query the nearest neighbor SMPL weight $\mathbf{W}_{smpl}=K(\mathbf{x},\mathbf{v})$
\begin{equation}
    \mathbf{x}_{cnl}^r = \sum _{j\,\in\, J} w_{j}\left(\left(\mathbf{x}_r\cdot R_j\right)+T_j\right)
\end{equation}
where $J$ denotes a set of joints, $w_{j}$ is the deformation weight of the $j$-th channel of $\mathbf{W}_{smpl}$ on the current sampled point, and $\mathbf{x}_{cnl}^r$ represents the coordinates of the sampled point in the canonical space after rigid deformation.

Directly using the nearest weight amounts to giving up non-rigid modeling of the possible clothing deformation caused by human body deformation, which often leads to unsatisfactory results. 
To enable the network to learn non-rigid deformation from limited images as much as possible, some methods~\cite{weng2022humanNeRF,peng2021neural,yu2023monohuman} introduce \emph{frame-level} features to facilitate learning. These features may include frame index or body pose, and for a particular frame, the frame-level features of all sampling points are generally consistent.
This method is useful in the task of synthesizing new viewpoints in dynamic scenes, but it is not suitable for \emph{animatable} human body. We designed a new network here to refine the coordinates of the sampled points in the canonical space. We use limited \emph{point-level} features to learn the offset of the sampling points in canonical space to improve the novel-view evaluation metric. The spatial-aware feature has shown significant advantages in this process because it aggregates vertex information within the receptive field of the sampling point in the current pose:
\begin{equation}
    \mathbf{x}_{cnl} = \mathbf{x}_{cnl}^r+P(\mathbf{F}_{s},\mathbf{x}_r)
\end{equation}
where $\mathbf{x}_{cnl}$ is the optimized result of the sampled point coordinates in the canonical space, and $P$ is the network module for optimizing the sampled point coordinates in the canonical space. 

\subsection{Appearance Net and Rendering}
\label{sec:AppearanceNetandRendering}

In order to better compare with methods such as HumanNeRF~\cite{weng2022humanNeRF}, which rely on neural networks to fit the deformation weights, we used the same neural radiance field structure. The coordinates of the sampled points in the canonical space were encoded using the same positional encoding as in NeRF, and the MLP was used to output the corresponding colors and densities:
\begin{equation}
    \mathbf{c},\sigma = M_{\sigma}(\mathbf{x}_{cnl})
\end{equation}
Finally, 
we render the neural human by the volume renderer \cite{max1995optical}. The rendered color $C(\mathbf{r})$ of the corresponding pixel with $N_s$ samples of ray $\mathbf{r}$ can be written as:
\begin{equation}
    \tilde{C}_{t}(\mathbf{r}) = \sum_{k=1}^{N_{s}} \left(\prod_{ j-1}^{ i -1}(1-\alpha_j)\right)\alpha _i\,\mathbf{c}(x_i)
\end{equation}
\begin{equation}
    \text{where}\, \, \alpha _i = 1-\exp(-\sigma \left(\mathbf{x}_i) \Delta\, t_i\right)
\end{equation}


\subsection{Training}

Given a set of monocular videos, the frame images of the video $\left \{ I_n|n=1,2,...,N \right \} $. Most of the images are used for training and the rest are used for evaluation. The foreground mask $\Theta_\mathrm{fore}$ is obtained from the density after the network output. For each foreground ray $\mathbf{r}\in \altmathcal{R}$, the corresponding loss function is defined as:
\begin{equation}
\label{lossall}
    \Loss  =  \Loss_{\mathrm{LPIPS}} + \lambda\Loss_{\mathrm{MSE}}
\end{equation}
\begin{equation}
\label{loss1}
    \text{where}\,\, \, \Loss_{\mathrm{MSE}} = \frac{1}{\left\|\altmathcal{R}\right\|} \sum_{\mathbf{r}\in \altmathcal{R}} \left\| \tilde{C}(\mathbf{r})-C(\mathbf{r})   \right\|_2^2
\end{equation}
\begin{equation}
\label{loss2}
   \text{and}\, \, \, \Loss_{\mathrm{LPIPS}} = lpips\left(\Theta_\mathrm{fore}(\tilde{I}_i ),\Theta_\mathrm{fore}(I_i)\right)
\end{equation}


\begin{figure*}[t]
    \centering
    
    \includegraphics[width=0.98\textwidth]{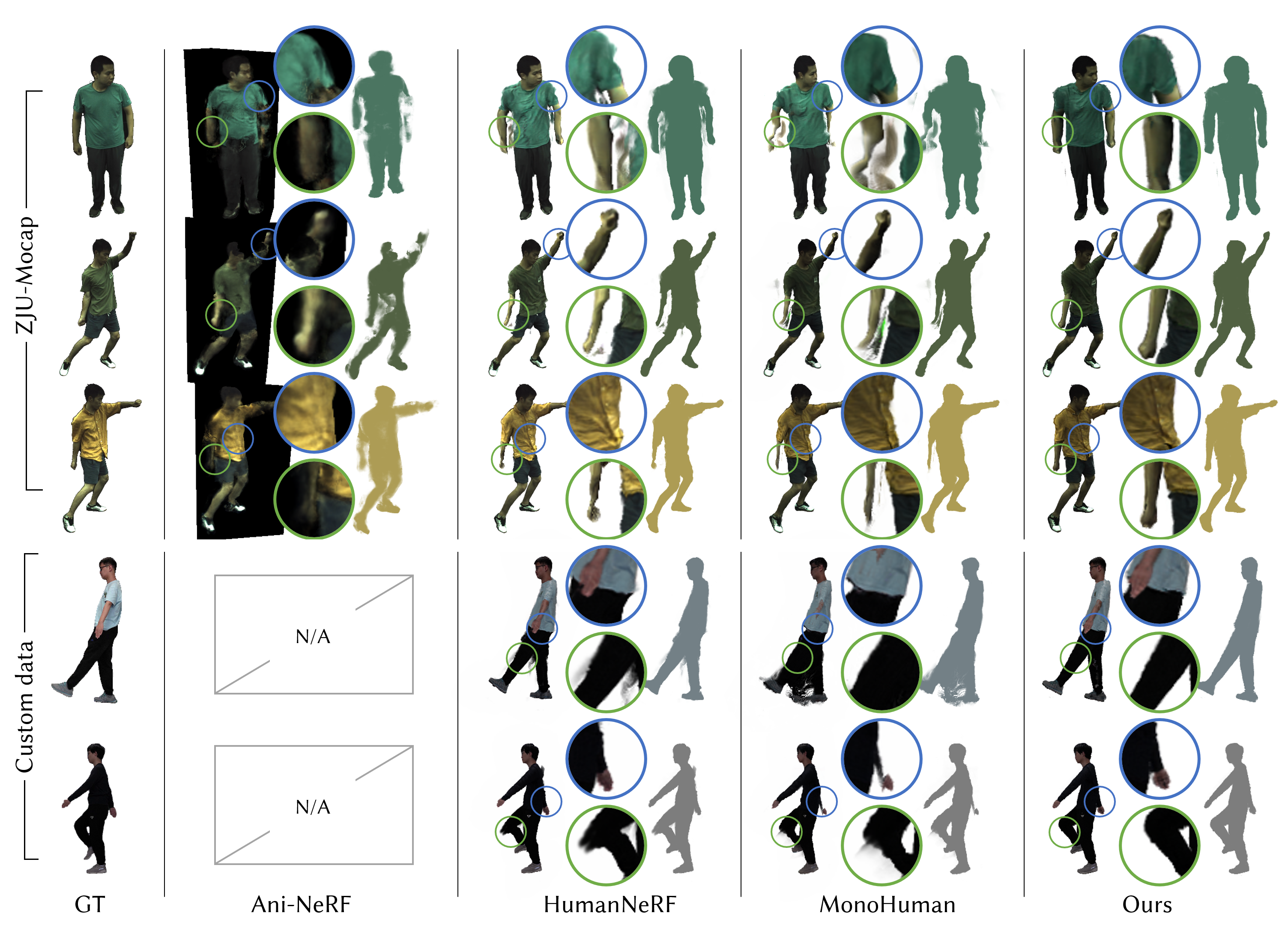}
      \vspace{-3mm}
    \caption{\textbf{Qualitative results with few-shot training images.} Because of limited information used in training, previous methods~\cite{weng2022humanNeRF,yu2023monohuman,peng2021animatable} cannot learn appropriate human weights. 
    The official code of Ani-NeRF~\cite{peng2021animatable} did not produce reasonable results on our data since it is designed for multi-camera input.
    HumanNeRF~\cite{weng2022humanNeRF} exhibits distortion and artifacts. The performance of Monohuman~\cite{yu2023monohuman}  is heavily influenced by the specific data.}
    \label{fig:quantitative}
\end{figure*}

\section{Experiments}

\begin{figure*}
    \centering
    \includegraphics[width=\textwidth]{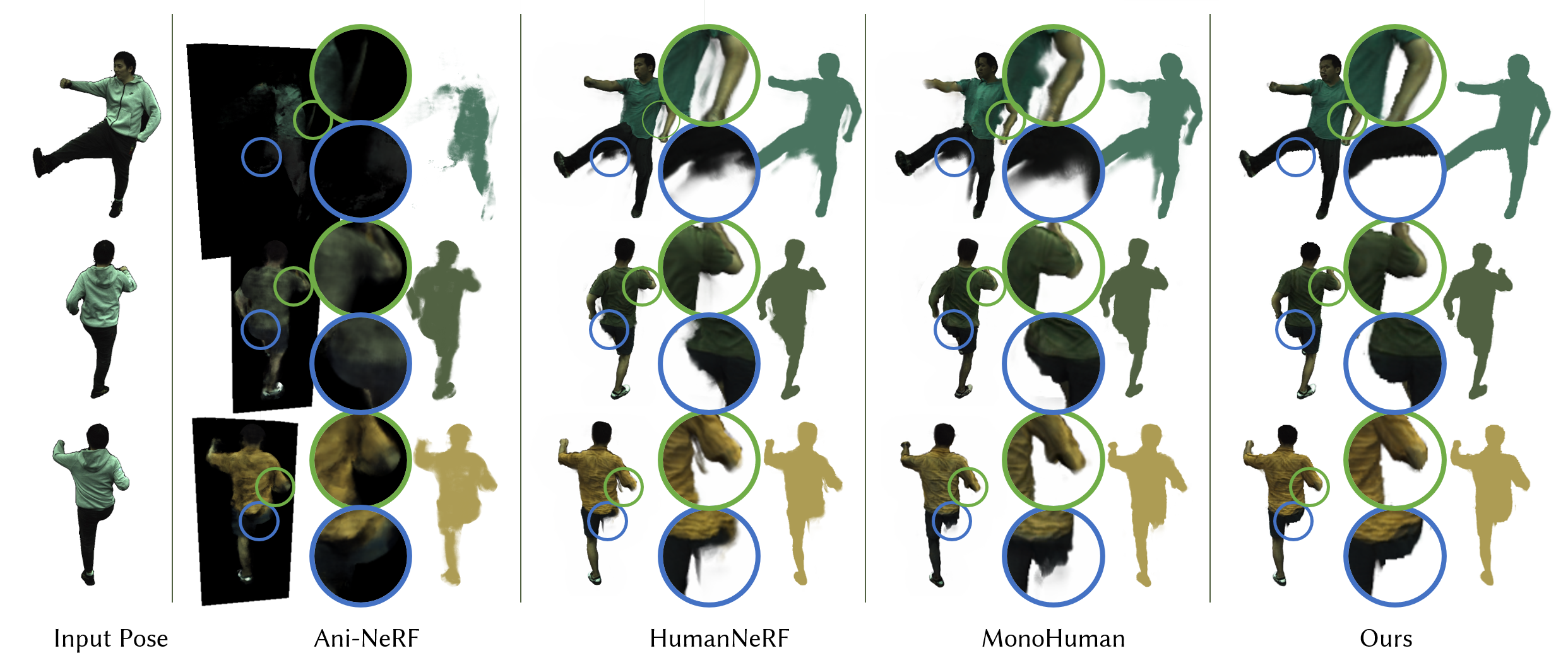}
    \vspace{-8mm}
    \caption{\textbf{Rendering results with pose sequences from Subject-387 in \textsc{ZJU-MoCap}.} 
    We use all the videos of the performers to train and synthesize images with different pose sequences from Subject-387. The baselines produce noticeable artifacts, while our method maintains high-quality image synthesis.
    }
    \label{fig:zju-nogt}
\end{figure*}

\begin{table*}[!t]
\centering
\caption{\textbf{Average results of six subjects on \textsc{ZJU-MoCap}}. Our method (blue) exhibits excellent quantitative metrics, especially in terms of LPIPS. This indicates that the results of our method are more in line with human visual perception. Our method demonstrated a better ability to avoid overfitting compared to other methods on our custom data. The official code of Ani-NeRF did not produce reasonable results on our custom data.
$^\dagger$LPIPS = 1,000$\times$LPIPS. 
}
\vspace{-3mm}
\label{tab:quantitive-results}
\setlength{\tabcolsep}{8pt}
\resizebox{\textwidth}{!}{
\begin{tabular}{lllllllllllll}
\toprule
\multicolumn{1}{c}{\multirow{3}{*}{Methods}}
&\multicolumn{6}{c}{\textsc{ZJU-MoCap}}
&\multicolumn{6}{c}{\textsc{In-the-wild Data}} \\
\cmidrule(lr){2-7}\cmidrule(lr){8-13}

& \multicolumn{3}{c}{{Full}} 
& \multicolumn{3}{c}{{Few-shot}} 
& \multicolumn{3}{c}{{Full}} 
& \multicolumn{3}{c}{{Few-shot}}  \\

\cmidrule(lr){2-4}\cmidrule(lr){5-7}
\cmidrule(lr){8-10}\cmidrule(lr){11-13}

 & \multicolumn{1}{c}{\small{PSNR$\uparrow$}} & \multicolumn{1}{c}{\small{SSIM$\uparrow$}} & \multicolumn{1}{c}{\small{$^\dagger$LPIPS$\downarrow$}} & \multicolumn{1}{c}{\small{PSNR$\uparrow$}} & \multicolumn{1}{c}{\small{SSIM$\uparrow$}} & \multicolumn{1}{c}{\small{$^\dagger$LPIPS$\downarrow$}} & \multicolumn{1}{c}{\small{PSNR$\uparrow$}} & \multicolumn{1}{c}{\small{SSIM$\uparrow$}} & \multicolumn{1}{c}{\small{$^\dagger$LPIPS$\downarrow$}} & \multicolumn{1}{c}{\small{PSNR$\uparrow$}} & \multicolumn{1}{c}{\small{SSIM$\uparrow$}} & \multicolumn{1}{c}{\small{$^\dagger$LPIPS$\downarrow$}} 

 \\ 
\midrule     
Ani-NeRF~\cite{peng2021animatable} &21.24 & 0.8458 & 68.221 &22.18 & 0.8339 & 64.839 & \multicolumn{1}{c}{--} & \multicolumn{1}{c}{--} & \multicolumn{1}{c}{--} & \multicolumn{1}{c}{--} & \multicolumn{1}{c}{--} & \multicolumn{1}{c}{--}\\

HumanNeRF~\cite{weng2022humanNeRF} & \textbf{31.15}  & 0.9739 & 24.822 & 29.90 & 0.9683 & 33.056 &28.97  &0.9629 &48.128 &28.82 &0.9618 &50.240 \\

MonoHuman~\cite{yu2023monohuman} & 30.91 & 0.9718 & 31.292 & 30.10 & 0.9677  & 36.494 &29.15 &0.9639 &51.623 &29.21	&0.9636	&56.220 \\

\rowcolor{CBLightGreen} Ours  & 31.09 & \textbf{0.9740} & \textbf{24.085} & \textbf{30.11} & \textbf{0.9684}  & \textbf{32.084}  &\textbf{29.23} & \textbf{0.9666} & \textbf{46.308} & \textbf{29.26} & \textbf{0.9669}  & \textbf{47.161} \\
 
\bottomrule
\end{tabular}
}
\vspace{-3mm}
\end{table*}

\paragraph{Evaluation Metrics.}

We use three metrics: peak signal-to-noise ratio (PSNR), structural similarity index (SSIM), and learned perceptual image patch similarity (LPIPS). 
It should be noted that LPIPS is the most \emph{human-perceptually-aligned} metric among these indicators, while PSNR prefers smooth results but may have bad visual quality~\cite{zhang2018unreasonable}.

\paragraph{Dataset.}
We use \textsc{ZJU-MoCap} and our captured in-the-wild videos to evaluate our method. We follow~\cite{weng2022humanNeRF}~\cite{yu2023monohuman}  to select the same six subjects in \textsc{ZJU-MoCap} for our evaluation.  
\textsc{ZJU-MoCap} is a dataset that captures the target human body from 23 different perspectives synchronously in a professional light stage room. We only use the first view captured in each subject.

The previous work did not consider the issue of pose leakage caused by the strong repetition of actions in \textsc{ZJU-MoCap}.
In order to further validate the performance of the model and make the task more applicable, we shot videos using handheld devices. We limited the training videos to a person spinning one round and used diverse action videos for evaluation, which is a more real-world applicable evaluation method.

\paragraph{Competed Methods.}
We compared our method in terms of the performance of image synthesis with the most influential method HumanNeRF~\cite{weng2022humanNeRF} and the latest state-of-the-art method MonoHuman~\cite{yu2023monohuman} which improves the pose generalization of HumanNeRF and works better than~\cite{jiang2022neuman,peng2021neural}. We also take Ani-NeRF~\cite{peng2021animatable} as one of the baselines because this work presents SMPL-based neural blend weight that can better generalize novel poses.


\subsection{Quantitative Evaluation}



For any set of data in \textsc{ZJU-MoCap}, we divided the data into training and testing data in a 4:1 ratio, which follows the setting of previous work. Differently, we uniformly select only about 30 frames from the divided training data as few-shot input. 
This setting can avoid pose leakage compared to the full input in a certain space.
Our experimental results in \textsc{ZJU-MoCap} with few-shot training images are shown in Figure~\ref{fig:quantitative}, and all the experimental results are shown in Table~\ref{tab:quantitive-results}. 

For our custom in-the-wild data, we train with a video of a performer spinning one round and use another video of the same person doing different poses for evaluation. The results are shown in Figure~\ref{fig:quantitative}.

\begin{table}[t]
    \centering
    \caption{
    \textbf{Comparison of training and rendering time.}
    Our method does not use existing acceleration modules. The simple yet effective architecture greatly reduces the computation required, thereby improving the overall speed.}
    \vspace{-3mm}
    \footnotesize
    \begin{tabular}{lrr}
    \toprule
         &Training &Rendering  \\
    \midrule
    Ani-NeRF~\cite{peng2021animatable}     &40 h   &2.73 s/it\\
    HumanNeRF~\cite{weng2022humanNeRF}     &56 h   &2.37 s/it\\
    MonoHuman~\cite{yu2023monohuman}       &70 h   &5.96 s/it\\
    \midrule
    Ours     & \textbf{2.5 h}     & \textbf{0.16 s/it} \\
    \bottomrule
    \end{tabular}
    \label{tab:time}
    
\end{table}

\subsection{Qualitative Evaluation}

In previous work, researchers often divide data into a certain ratio as novel pose experiments for quantitative evaluation.
However, in the most widely used dataset \textsc{ZJU-MoCap}, performers repeat the same action in a set of data. 
This leads to the poses in evaluation data being highly similar to poses used for training.
Although we mitigate pose leakage issues through few-shot input, a considerable portion of poses in the evaluation data are still similar, resulting in limited difference in the average results.

\begin{figure}[!t]
  \centering
  \begin{subfigure}[b]{0.4\linewidth}
    \includegraphics[width=\linewidth]{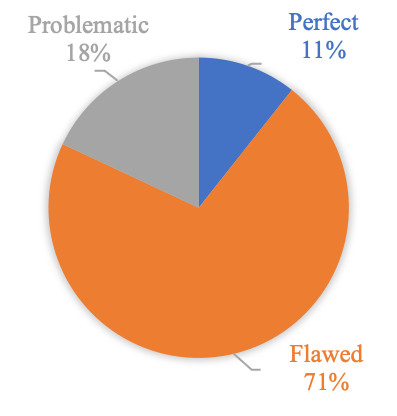}
    \caption{HumanNeRF~\cite{weng2022humanNeRF}}
    \label{fig:humanNeRF-sub}
  \end{subfigure}
  \quad
  \begin{subfigure}[b]{0.4\linewidth}
    \includegraphics[width=\linewidth]{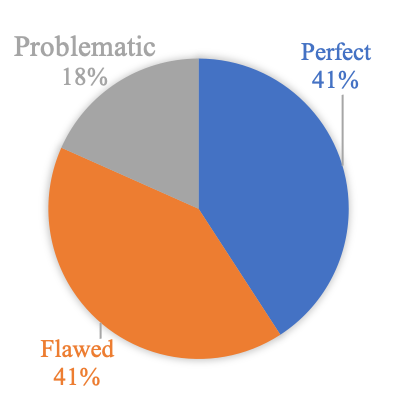}
    \caption{Ours}
    \label{fig:ours-sub}
  \end{subfigure}
  \vspace{-3mm}
  \caption{\textbf{Subjective evaluation results.} We use cross-rendering of different subjects in the \textsc{ZJU-MoCap} to ensure that the evaluation of the novel pose is sufficiently novel. Our results show a significantly higher perfect proportion than HumaneNeRF.}
  \label{fig:subjective-reasearch}
\end{figure}

A simple way to avoid pose leakage problem is to use completely different action sequences as evaluation. 
But as there is no ground truth of \textsc{ZJU-MoCap}, we provided comparison results as Figure~\ref{fig:zju-nogt}, and synthesized over 200 zero-shot pose images and converted them into 20 videos, which were subjectively evaluated by six or more participants as shown in Figure~\ref{fig:subjective-reasearch}.

The entire evaluation process is single-blind, meaning that the participants do not know which specific method generated the results. We also included some test seeds, which serve as the Ground Truth, and all of these seeds received high scores from the participants, indicating that their evaluations were professional and objective. The participants were asked to evaluate the video in terms of image distortion, artifacts, details, plausibility, and precision, and to provide a final score that was assigned to one of three different levels.

As a result, almost all the participants subjectively think our results have a better performance. Our results were classified as perfect in significantly higher numbers than HumanNeRF, with fewer votes for flaws. However, both the results performed poorly in cases of poor data quality.

\subsection{Ablation Studies}

\begin{figure}[t]
  \centering
  \begin{subfigure}[b]{0.44\linewidth}
    \includegraphics[width=\linewidth]{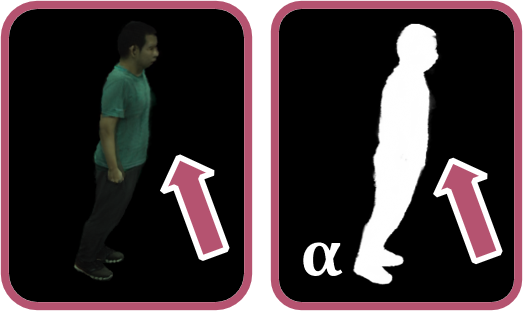}
    \caption*{\emph{w/} Conv-Filter}
    \label{fig:wconfilter}
  \end{subfigure}
  \quad
  \begin{subfigure}[b]{0.44\linewidth}
    \includegraphics[width=\linewidth]{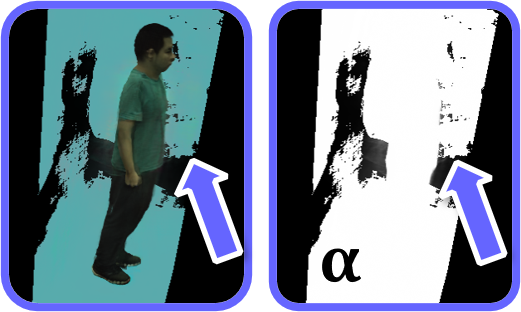}
    \caption*{\emph{w/o} Conv-Filter}
    \label{fig:w0confilter}
  \end{subfigure}
  \vspace{-3mm}
    \caption{\textbf{Ablation study on Conv-Filter.}}
  \label{fig:ablation-convfilter}
\end{figure}

\paragraph{Conv-Filter.}
In our method, the point filter is essential. Our experiments showed that if the sampled points are not filtered, we cannot learn the correct alpha map (see Figure~\ref{fig:ablation-convfilter}) in our experiments, and it greatly increases the computational complexity, requiring longer training time. 
We further investigated the reason why the phenomenon of alpha map learning errors occurs due to color diffusion into the surrounding space, and we believe that this is determined by the distribution of skin weight. As shown in Figure~\ref{fig:motivation}, the learnable weights tend to give negative values to irrelevant joint weights, but this is unreasonable. It can be explained by the fact that these methods do not require filters to avoid the phenomenon, because we observe similar phenomena when we map the learnable weights to the same distribution as ours through sigmoid.


\begin{figure}[t]
  \centering
  \begin{subfigure}[b]{0.44\linewidth}
    \includegraphics[width=\linewidth]{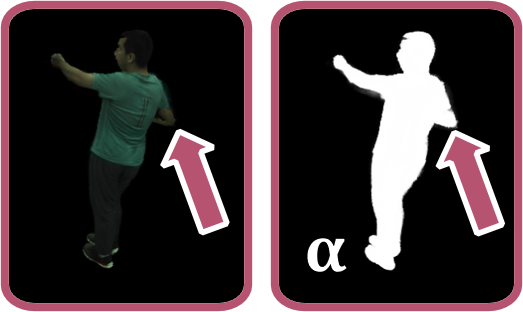}
    \caption*{\emph{w/} Point Refinement}
    \label{fig:wpointrefine}
  \end{subfigure}
  \quad
  \begin{subfigure}[b]{0.44\linewidth}
    \includegraphics[width=\linewidth]{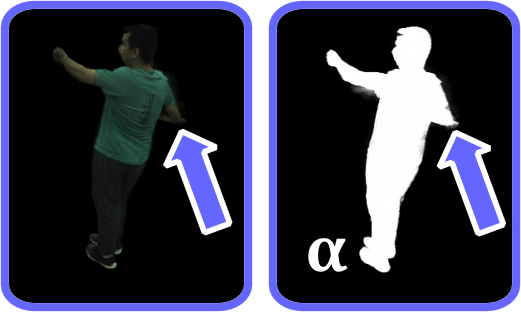}
    \caption*{\emph{w/o} Point Refinement}
    \label{fig:wopointrefine}
  \end{subfigure}
  \vspace{-3mm}
  \caption{\textbf{Ablation study on Point Refinement.}}
  \label{fig:pointrefiner}
\end{figure}

\begin{figure}[t]
  \centering
  \begin{subfigure}[b]{0.44\linewidth}
    \includegraphics[width=\linewidth]{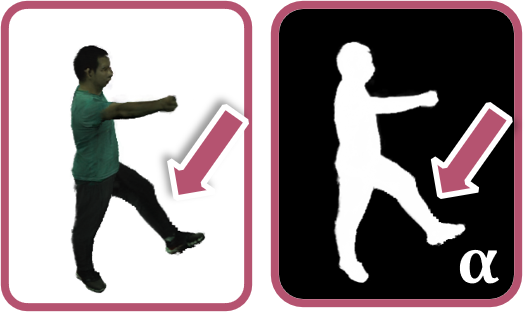}
    \caption*{Point-level feature (Ours)}
    \label{fig:plf}
  \end{subfigure}
  \quad
\begin{subfigure}[b]{0.44\linewidth}
    \includegraphics[width=\linewidth]{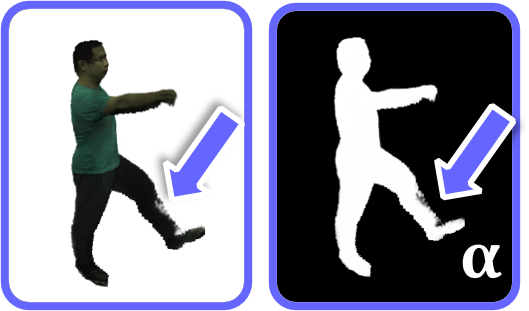}
    \caption*{Frame-level feature}
    \label{fig:flf}
  \end{subfigure}
  \vspace{-3mm}
  \caption{\textbf{Ablation study on the input of Point Refinement}.
  }
  \label{fig:framefeature}
\end{figure}

\paragraph{Point-level Feature Refine.}
It is a common practice to add offsets to the deformation process using neural networks, but previous methods often used time or the pose of the current frame as control information. 
This frame-level feature often leads to overfitting, but in previous experiments, this phenomenon was not significant due to the similarity and repetition of actions in \textsc{ZJU-MoCap}.
We extract point-level spatial-aware feature while filtering points in ConvFilter. This not only corrects the unnatural joints caused by rigid deformation (see Figure~\ref{fig:pointrefiner}, Table~\ref{table:ablationtable}) but also avoids overfitting compared to the previous frame-level feature (see Figure~\ref{fig:framefeature}).




\begin{table}[t]
\centering
\caption{\textbf{Quantitative ablation study in \textsc{ZJU-MoCap}.} 
We evaluate the effectiveness of canonical points refined with $F_s$ and Conv-Filter.
}
\vspace{-3mm}
\footnotesize
{
\begin{tabular}{llll}
\toprule
 & \multicolumn{1}{c}{{PSNR$\uparrow$}} & \multicolumn{1}{c}{{SSIM$\uparrow$}} & \multicolumn{1}{c}{$^\dagger$LPIPS$\downarrow$} \\ 
\midrule
 \emph{w/o} $F_{s}$ refine & 30.93 & 0.971 & ~~24.712 \\ 
\emph{w/o} filter & ~~9.22   &0.607   &318.460   \\
\midrule
 Full & \textbf{31.09} & \textbf{0.974} & ~~\textbf{24.085} \\
 \bottomrule
\end{tabular}
\vspace{-4mm}
}
\label{table:ablationtable}
\end{table}


\section{Limitations}
Our method is state-of-the-art in the task of learning implicit human representations from limited input and synthesizing diverse pose images. 
What's required is only a monocular video, even a few images, and easily obtainable SMPL information, without the need for additional calculations of texture information, greatly expanding the method's universality. 
However, our method still has certain limitations: 
1) The effectiveness of our method depends on the accuracy of the estimated SMPL, and when the SMPL accuracy is low, the results may be blurred. Currently, the accuracy of SMPL estimation methods is not always satisfactory.
2) Our method uses coordinate voxelization to assist calculation, which may cause image edge serrations. Fine-tuning can be achieved by adjusting the voxel size and convolution kernel size, which will increase computational cost.
3) Our method uses basic SMPL information for training, so it is difficult to drive hand and facial details.

\section{Conclusion}
We propose a human neural radiance field model that can train with limited inputs and generalize to diverse zero-shot poses. 
Unlike previous methods, our approach filters the sampling points and obtains point-level features in a voxel volume of explicit vertices, and subsequently deforms the points to a canonical space using general and specific mapping. 
Our approach uses less than 1\% learnable parameters and achieves state-of-the-art novel pose metrics in our experiments, and maintains the best performance with few-shot input.
Our approach only requires estimated SMPL information, which can be easily obtained using existing methods, thereby maintaining usability while being able to generalize to industries such as video production.

\vspace{2mm}
{\small
\heading{Acknowledgements}.
This work was supported by the Hubei Key R\&D Project (2022BAA033), the National Natural Science Foundation of China (62171325), the Beijing Nova Program (20220484063), and the Value Exchange Engineering, a joint research project between Mercari, Inc. and RIISE. The numerical calculations in this paper have been done on the supercomputing system in the Supercomputing Center of Wuhan University.
}

{
    \small
    \bibliographystyle{ieeenat_fullname}
    \bibliography{main}

\begin{thebibliography}{70}
\providecommand{\natexlab}[1]{#1}
\providecommand{\url}[1]{\texttt{#1}}
\expandafter\ifx\csname urlstyle\endcsname\relax
  \providecommand{\doi}[1]{doi: #1}\else
  \providecommand{\doi}{doi: \begingroup \urlstyle{rm}\Url}\fi

\bibitem[Aliev et~al.(2020)Aliev, Sevastopolsky, Kolos, Ulyanov, and
  Lempitsky]{aliev2020neural}
Kara-Ali Aliev, Artem Sevastopolsky, Maria Kolos, Dmitry Ulyanov, and Victor
  Lempitsky.
\newblock Neural point-based graphics.
\newblock In \emph{Proceedings of the European Conference on Computer Vision},
  2020.

\bibitem[Barron et~al.(2021)Barron, Mildenhall, Tancik, Hedman, Martin-Brualla,
  and Srinivasan]{barron2021mip}
Jonathan~T Barron, Ben Mildenhall, Matthew Tancik, Peter Hedman, Ricardo
  Martin-Brualla, and Pratul~P Srinivasan.
\newblock Mip-{N}e{R}{F}: A multiscale representation for anti-aliasing neural
  radiance fields.
\newblock In \emph{Proceedings of the IEEE/CVF International Conference on
  Computer Vision}, pages 5855--5864, 2021.

\bibitem[Chan et~al.(2022)Chan, Lin, Chan, Nagano, Pan, De~Mello, Gallo,
  Guibas, Tremblay, Khamis, et~al.]{chan2022efficient}
Eric~R Chan, Connor~Z Lin, Matthew~A Chan, Koki Nagano, Boxiao Pan, Shalini
  De~Mello, Orazio Gallo, Leonidas~J Guibas, Jonathan Tremblay, Sameh Khamis,
  et~al.
\newblock Efficient geometry-aware 3{D} generative adversarial networks.
\newblock In \emph{Proceedings of the IEEE/CVF Conference on Computer Vision
  and Pattern Recognition}, 2022.

\bibitem[Chen et~al.(2022)Chen, Zhang, Xu, Liu, Cai, Feng, and
  Yan]{chen2022geometry}
Mingfei Chen, Jianfeng Zhang, Xiangyu Xu, Lijuan Liu, Yujun Cai, Jiashi Feng,
  and Shuicheng Yan.
\newblock Geometry-guided progressive {N}e{R}{F} for generalizable and
  efficient neural human rendering.
\newblock In \emph{Proceedings of the European Conference on Computer Vision},
  2022.

\bibitem[Chen et~al.(2023)Chen, Wang, Chen, Zhang, Li, Guo, Wang, and
  Wang]{chen2023uv}
Yue Chen, Xuan Wang, Xingyu Chen, Qi Zhang, Xiaoyu Li, Yu Guo, Jue Wang, and
  Fei Wang.
\newblock Uv volumes for real-time rendering of editable free-view human
  performance.
\newblock In \emph{Proceedings of the IEEE/CVF Conference on Computer Vision
  and Pattern Recognition}, 2023.

\bibitem[Cheng et~al.(2022)Cheng, Xu, Piao, Qian, Wu, Lin, and
  Li]{cheng2022generalizable}
Wei Cheng, Su Xu, Jingtan Piao, Chen Qian, Wayne Wu, Kwan-Yee Lin, and
  Hongsheng Li.
\newblock Generalizable neural performer: Learning robust radiance fields for
  human novel view synthesis.
\newblock \emph{arXiv preprint arXiv:2204.11798}, 2022.

\bibitem[Contributors(2022)]{spconv2022}
SpConv Contributors.
\newblock Spconv: Spatially sparse convolution library.
\newblock \url{https://github.com/traveller59/spconv}, 2022.

\bibitem[Flynn et~al.(2019)Flynn, Broxton, Debevec, DuVall, Fyffe, Overbeck,
  Snavely, and Tucker]{flynn2019deepview}
John Flynn, Michael Broxton, Paul Debevec, Matthew DuVall, Graham Fyffe, Ryan
  Overbeck, Noah Snavely, and Richard Tucker.
\newblock Deepview: View synthesis with learned gradient descent.
\newblock In \emph{Proceedings of the IEEE/CVF Conference on Computer Vision
  and Pattern Recognition}, 2019.

\bibitem[Gao et~al.(2021)Gao, Saraf, Kopf, and Huang]{gao2021dynamic}
Chen Gao, Ayush Saraf, Johannes Kopf, and Jia-Bin Huang.
\newblock Dynamic view synthesis from dynamic monocular video.
\newblock In \emph{Proceedings of the IEEE/CVF International Conference on
  Computer Vision}, 2021.

\bibitem[Gao et~al.(2022)Gao, Yang, Kim, Peng, Liu, and Tong]{gao2022mpsNeRF}
Xiangjun Gao, Jiaolong Yang, Jongyoo Kim, Sida Peng, Zicheng Liu, and Xin Tong.
\newblock Mps-{N}e{R}{F}: Generalizable 3{D} human rendering from multiview
  images.
\newblock \emph{IEEE Transactions on Pattern Analysis and Machine
  Intelligence}, 2022.

\bibitem[Geng et~al.(2023)Geng, Peng, Xu, Bao, and Zhou]{geng2023learning}
Chen Geng, Sida Peng, Zhen Xu, Hujun Bao, and Xiaowei Zhou.
\newblock Learning neural volumetric representations of dynamic humans in
  minutes.
\newblock \emph{arXiv preprint arXiv:2302.12237}, 2023.

\bibitem[Habermann et~al.(2020)Habermann, Xu, Zollhofer, Pons-Moll, and
  Theobalt]{habermann2020deepcap}
Marc Habermann, Weipeng Xu, Michael Zollhofer, Gerard Pons-Moll, and Christian
  Theobalt.
\newblock Deepcap: Monocular human performance capture using weak supervision.
\newblock In \emph{Proceedings of the IEEE/CVF Conference on Computer Vision
  and Pattern Recognition}, 2020.

\bibitem[Hedman et~al.(2021)Hedman, Srinivasan, Mildenhall, Barron, and
  Debevec]{hedman2021baking}
Peter Hedman, Pratul~P Srinivasan, Ben Mildenhall, Jonathan~T Barron, and Paul
  Debevec.
\newblock Baking neural radiance fields for real-time view synthesis.
\newblock In \emph{Proceedings of the IEEE/CVF International Conference on
  Computer Vision}, 2021.

\bibitem[Jayasundara et~al.(2023)Jayasundara, Agrawal, Heron, Shrivastava, and
  Davis]{jayasundara2023flexNeRF}
Vinoj Jayasundara, Amit Agrawal, Nicolas Heron, Abhinav Shrivastava, and
  Larry~S Davis.
\newblock Flex{N}e{R}{F}: Photorealistic free-viewpoint rendering of moving
  humans from sparse views.
\newblock In \emph{Proceedings of the IEEE/CVF Conference on Computer Vision
  and Pattern Recognition}, 2023.

\bibitem[Jiang et~al.(2022{\natexlab{a}})Jiang, Hong, Bao, and
  Zhang]{jiang2022selfrecon}
Boyi Jiang, Yang Hong, Hujun Bao, and Juyong Zhang.
\newblock Selfrecon: Self reconstruction your digital avatar from monocular
  video.
\newblock In \emph{Proceedings of the IEEE/CVF Conference on Computer Vision
  and Pattern Recognition}, 2022{\natexlab{a}}.

\bibitem[Jiang et~al.(2023)Jiang, Chen, Song, and
  Hilliges]{jiang2023instantavatar}
Tianjian Jiang, Xu Chen, Jie Song, and Otmar Hilliges.
\newblock Instantavatar: Learning avatars from monocular video in 60 seconds.
\newblock In \emph{Proceedings of the IEEE/CVF Conference on Computer Vision
  and Pattern Recognition}, 2023.

\bibitem[Jiang et~al.(2022{\natexlab{b}})Jiang, Yi, Samei, Tuzel, and
  Ranjan]{jiang2022neuman}
Wei Jiang, Kwang~Moo Yi, Golnoosh Samei, Oncel Tuzel, and Anurag Ranjan.
\newblock Neuman: Neural human radiance field from a single video.
\newblock In \emph{Proceedings of the European Conference on Computer Vision},
  2022{\natexlab{b}}.

\bibitem[Kwon et~al.(2021)Kwon, Kim, Ceylan, and Fuchs]{kwon2021neural}
Youngjoong Kwon, Dahun Kim, Duygu Ceylan, and Henry Fuchs.
\newblock Neural human performer: Learning generalizable radiance fields for
  human performance rendering.
\newblock \emph{Advances in Neural Information Processing Systems},
  34:\penalty0 24741--24752, 2021.

\bibitem[Li et~al.(2024)Li, Qian, and Xia]{li2024unleashing}
Guopeng Li, Ming Qian, and Gui-Song Xia.
\newblock Unleashing unlabeled data: A paradigm for cross-view
  geo-localization.
\newblock \emph{arXiv preprint arXiv:2403.14198}, 2024.

\bibitem[Li et~al.(2022)Li, Tanke, Vo, Zollh{\"o}fer, Gall, Kanazawa, and
  Lassner]{li2022tava}
Ruilong Li, Julian Tanke, Minh Vo, Michael Zollh{\"o}fer, J{\"u}rgen Gall,
  Angjoo Kanazawa, and Christoph Lassner.
\newblock Tava: Template-free animatable volumetric actors.
\newblock In \emph{Proceedings of the European Conference on Computer Vision},
  2022.

\bibitem[Li et~al.(2021)Li, Niklaus, Snavely, and Wang]{li2021neural}
Zhengqi Li, Simon Niklaus, Noah Snavely, and Oliver Wang.
\newblock Neural scene flow fields for space-time view synthesis of dynamic
  scenes.
\newblock In \emph{Proceedings of the IEEE/CVF Conference on Computer Vision
  and Pattern Recognition}, 2021.

\bibitem[Li et~al.(2023)Li, Zheng, Liu, Zhou, and Liu]{li2023posevocab}
Zhe Li, Zerong Zheng, Yuxiao Liu, Boyao Zhou, and Yebin Liu.
\newblock Posevocab: Learning joint-structured pose embeddings for human avatar
  modeling.
\newblock \emph{arXiv preprint arXiv:2304.13006}, 2023.

\bibitem[Liao et~al.(2020)Liao, Schwarz, Mescheder, and
  Geiger]{liao2020towards}
Yiyi Liao, Katja Schwarz, Lars Mescheder, and Andreas Geiger.
\newblock Towards unsupervised learning of generative models for 3{D}
  controllable image synthesis.
\newblock In \emph{Proceedings of the IEEE/CVF conference on computer vision
  and pattern recognition}, 2020.

\bibitem[Liu et~al.(2019)Liu, Xu, Zollhoefer, Kim, Bernard, Habermann, Wang,
  and Theobalt]{liu2019neural}
Lingjie Liu, Weipeng Xu, Michael Zollhoefer, Hyeongwoo Kim, Florian Bernard,
  Marc Habermann, Wenping Wang, and Christian Theobalt.
\newblock Neural rendering and reenactment of human actor videos.
\newblock \emph{ACM Transactions on Graphics (TOG)}, 38\penalty0 (5):\penalty0
  1--14, 2019.

\bibitem[Liu et~al.(2020{\natexlab{a}})Liu, Gu, Zaw~Lin, Chua, and
  Theobalt]{liu2020neural}
Lingjie Liu, Jiatao Gu, Kyaw Zaw~Lin, Tat-Seng Chua, and Christian Theobalt.
\newblock Neural sparse voxel fields.
\newblock \emph{Advances in Neural Information Processing Systems},
  33:\penalty0 15651--15663, 2020{\natexlab{a}}.

\bibitem[Liu et~al.(2021)Liu, Habermann, Rudnev, Sarkar, Gu, and
  Theobalt]{liu2021neural}
Lingjie Liu, Marc Habermann, Viktor Rudnev, Kripasindhu Sarkar, Jiatao Gu, and
  Christian Theobalt.
\newblock Neural actor: Neural free-view synthesis of human actors with pose
  control.
\newblock \emph{ACM Transactions on Graphics (TOG)}, 40\penalty0 (6):\penalty0
  1--16, 2021.

\bibitem[Liu et~al.(2020{\natexlab{b}})Liu, Zhang, Peng, Shi, Pollefeys, and
  Cui]{liu2020dist}
Shaohui Liu, Yinda Zhang, Songyou Peng, Boxin Shi, Marc Pollefeys, and Zhaopeng
  Cui.
\newblock Dist: Rendering deep implicit signed distance function with
  differentiable sphere tracing.
\newblock In \emph{Proceedings of the IEEE/CVF Conference on Computer Vision
  and Pattern Recognition}, 2020{\natexlab{b}}.

\bibitem[Lombardi et~al.(2019)Lombardi, Simon, Saragih, Schwartz, Lehrmann, and
  Sheikh]{lombardi2019neural}
Stephen Lombardi, Tomas Simon, Jason Saragih, Gabriel Schwartz, Andreas
  Lehrmann, and Yaser Sheikh.
\newblock Neural volumes: Learning dynamic renderable volumes from images.
\newblock \emph{arXiv preprint arXiv:1906.07751}, 2019.

\bibitem[Loper et~al.(2015)Loper, Mahmood, Romero, Pons-Moll, and
  Black]{loper2015smpl}
Matthew Loper, Naureen Mahmood, Javier Romero, Gerard Pons-Moll, and Michael~J
  Black.
\newblock {S}{M}{P}{L}: A skinned multi-person linear model.
\newblock \emph{ACM transactions on graphics (TOG)}, 34\penalty0 (6):\penalty0
  1--16, 2015.

\bibitem[Martin-Brualla et~al.(2018)Martin-Brualla, Pandey, Yang, Pidlypenskyi,
  Taylor, Valentin, Khamis, Davidson, Tkach, Lincoln,
  et~al.]{martin2018lookingood}
Ricardo Martin-Brualla, Rohit Pandey, Shuoran Yang, Pavel Pidlypenskyi,
  Jonathan Taylor, Julien Valentin, Sameh Khamis, Philip Davidson, Anastasia
  Tkach, Peter Lincoln, et~al.
\newblock Lookingood: Enhancing performance capture with real-time neural
  re-rendering.
\newblock \emph{arXiv preprint arXiv:1811.05029}, 2018.

\bibitem[Martin-Brualla et~al.(2021)Martin-Brualla, Radwan, Sajjadi, Barron,
  Dosovitskiy, and Duckworth]{martin2021NeRF}
Ricardo Martin-Brualla, Noha Radwan, Mehdi~SM Sajjadi, Jonathan~T Barron,
  Alexey Dosovitskiy, and Daniel Duckworth.
\newblock {N}e{R}{F} in the wild: Neural radiance fields for unconstrained
  photo collections.
\newblock In \emph{Proceedings of the IEEE/CVF Conference on Computer Vision
  and Pattern Recognition}, 2021.

\bibitem[Max(1995)]{max1995optical}
Nelson Max.
\newblock Optical models for direct volume rendering.
\newblock \emph{IEEE Transactions on Visualization and Computer Graphics},
  1\penalty0 (2):\penalty0 99--108, 1995.

\bibitem[Mildenhall et~al.(2020)Mildenhall, Srinivasan, Tancik, Barron,
  Ramamoorthi, and Ng]{mildenhall2021NeRF}
Ben Mildenhall, Pratul~P. Srinivasan, Matthew Tancik, Jonathan~T. Barron, Ravi
  Ramamoorthi, and Ren Ng.
\newblock {N}e{R}{F}: Representing scenes as neural radiance fields for view
  synthesis.
\newblock In \emph{Proceedings of the European Conference on Computer Vision},
  2020.

\bibitem[M{\"u}ller et~al.(2022)M{\"u}ller, Evans, Schied, and
  Keller]{muller2022instant}
Thomas M{\"u}ller, Alex Evans, Christoph Schied, and Alexander Keller.
\newblock Instant neural graphics primitives with a multiresolution hash
  encoding.
\newblock \emph{ACM Transactions on Graphics (ToG)}, 41\penalty0 (4):\penalty0
  1--15, 2022.

\bibitem[Niemeyer and Geiger(2021)]{niemeyer2021giraffe}
Michael Niemeyer and Andreas Geiger.
\newblock Giraffe: Representing scenes as compositional generative neural
  feature fields.
\newblock In \emph{Proceedings of the IEEE/CVF Conference on Computer Vision
  and Pattern Recognition}, 2021.

\bibitem[Niemeyer et~al.(2020)Niemeyer, Mescheder, Oechsle, and
  Geiger]{niemeyer2020differentiable}
Michael Niemeyer, Lars Mescheder, Michael Oechsle, and Andreas Geiger.
\newblock Differentiable volumetric rendering: Learning implicit 3{D}
  representations without 3{D} supervision.
\newblock In \emph{Proceedings of the IEEE/CVF Conference on Computer Vision
  and Pattern Recognition}, 2020.

\bibitem[Park et~al.(2021{\natexlab{a}})Park, Sinha, Barron, Bouaziz, Goldman,
  Seitz, and Martin-Brualla]{park2021NeRFies}
Keunhong Park, Utkarsh Sinha, Jonathan~T Barron, Sofien Bouaziz, Dan~B Goldman,
  Steven~M Seitz, and Ricardo Martin-Brualla.
\newblock {N}e{R}{F}ies: Deformable neural radiance fields.
\newblock In \emph{Proceedings of the IEEE/CVF International Conference on
  Computer Vision}, 2021{\natexlab{a}}.

\bibitem[Park et~al.(2021{\natexlab{b}})Park, Sinha, Hedman, Barron, Bouaziz,
  Goldman, Martin-Brualla, and Seitz]{park2021hyperNeRF}
Keunhong Park, Utkarsh Sinha, Peter Hedman, Jonathan~T Barron, Sofien Bouaziz,
  Dan~B Goldman, Ricardo Martin-Brualla, and Steven~M Seitz.
\newblock Hyper{N}e{R}{F}: A higher-dimensional representation for
  topologically varying neural radiance fields.
\newblock \emph{arXiv preprint arXiv:2106.13228}, 2021{\natexlab{b}}.

\bibitem[Peng et~al.(2023)Peng, Hu, Zhou, Gao, and Zhang]{peng2023intrinsicngp}
Bo Peng, Jun Hu, Jingtao Zhou, Xuan Gao, and Juyong Zhang.
\newblock Intrinsicngp: Intrinsic coordinate based hash encoding for human
  {N}e{R}{F}.
\newblock \emph{arXiv preprint arXiv:2302.14683}, 2023.

\bibitem[Peng et~al.(2021{\natexlab{a}})Peng, Dong, Wang, Zhang, Shuai, Zhou,
  and Bao]{peng2021animatable}
Sida Peng, Junting Dong, Qianqian Wang, Shangzhan Zhang, Qing Shuai, Xiaowei
  Zhou, and Hujun Bao.
\newblock Animatable neural radiance fields for modeling dynamic human bodies.
\newblock In \emph{Proceedings of the IEEE/CVF International Conference on
  Computer Vision}, 2021{\natexlab{a}}.

\bibitem[Peng et~al.(2021{\natexlab{b}})Peng, Zhang, Xu, Wang, Shuai, Bao, and
  Zhou]{peng2021neural}
Sida Peng, Yuanqing Zhang, Yinghao Xu, Qianqian Wang, Qing Shuai, Hujun Bao,
  and Xiaowei Zhou.
\newblock Neural body: Implicit neural representations with structured latent
  codes for novel view synthesis of dynamic humans.
\newblock In \emph{Proceedings of the IEEE/CVF Conference on Computer Vision
  and Pattern Recognition}, 2021{\natexlab{b}}.

\bibitem[Pumarola et~al.(2021)Pumarola, Corona, Pons-Moll, and
  Moreno-Noguer]{pumarola2021d}
Albert Pumarola, Enric Corona, Gerard Pons-Moll, and Francesc Moreno-Noguer.
\newblock D-{N}e{R}{F}: Neural radiance fields for dynamic scenes.
\newblock In \emph{Proceedings of the IEEE/CVF Conference on Computer Vision
  and Pattern Recognition}, 2021.

\bibitem[Saito et~al.(2019)Saito, Huang, Natsume, Morishima, Kanazawa, and
  Li]{saito2019pifu}
Shunsuke Saito, Zeng Huang, Ryota Natsume, Shigeo Morishima, Angjoo Kanazawa,
  and Hao Li.
\newblock Pifu: Pixel-aligned implicit function for high-resolution clothed
  human digitization.
\newblock In \emph{Proceedings of the IEEE/CVF international conference on
  computer vision}, 2019.

\bibitem[Saito et~al.(2020)Saito, Simon, Saragih, and Joo]{saito2020pifuhd}
Shunsuke Saito, Tomas Simon, Jason Saragih, and Hanbyul Joo.
\newblock Pifuhd: Multi-level pixel-aligned implicit function for
  high-resolution 3{D} human digitization.
\newblock In \emph{Proceedings of the IEEE/CVF Conference on Computer Vision
  and Pattern Recognition}, 2020.

\bibitem[Sitzmann et~al.(2019{\natexlab{a}})Sitzmann, Thies, Heide,
  Nie{\ss}ner, Wetzstein, and Zollhofer]{sitzmann2019deepvoxels}
Vincent Sitzmann, Justus Thies, Felix Heide, Matthias Nie{\ss}ner, Gordon
  Wetzstein, and Michael Zollhofer.
\newblock Deepvoxels: Learning persistent 3{D} feature embeddings.
\newblock In \emph{Proceedings of the IEEE/CVF Conference on Computer Vision
  and Pattern Recognition}, 2019{\natexlab{a}}.

\bibitem[Sitzmann et~al.(2019{\natexlab{b}})Sitzmann, Zollh{\"o}fer, and
  Wetzstein]{sitzmann2019scene}
Vincent Sitzmann, Michael Zollh{\"o}fer, and Gordon Wetzstein.
\newblock Scene representation networks: Continuous 3{D}-structure-aware neural
  scene representations.
\newblock \emph{Advances in Neural Information Processing Systems}, 32,
  2019{\natexlab{b}}.

\bibitem[Srinivasan et~al.(2021)Srinivasan, Deng, Zhang, Tancik, Mildenhall,
  and Barron]{srinivasan2021nerv}
Pratul~P Srinivasan, Boyang Deng, Xiuming Zhang, Matthew Tancik, Ben
  Mildenhall, and Jonathan~T Barron.
\newblock Nerv: Neural reflectance and visibility fields for relighting and
  view synthesis.
\newblock In \emph{Proceedings of the IEEE/CVF Conference on Computer Vision
  and Pattern Recognition}, 2021.

\bibitem[Su et~al.(2021)Su, Yu, Zollh{\"o}fer, and Rhodin]{su2021aNeRF}
Shih-Yang Su, Frank Yu, Michael Zollh{\"o}fer, and Helge Rhodin.
\newblock A-{N}e{R}{F}: Articulated neural radiance fields for learning human
  shape, appearance, and pose.
\newblock \emph{Advances in Neural Information Processing Systems},
  34:\penalty0 12278--12291, 2021.

\bibitem[Su et~al.(2023)Su, Bagautdinov, and Rhodin]{su2023npc}
Shih-Yang Su, Timur Bagautdinov, and Helge Rhodin.
\newblock Npc: Neural point characters from video.
\newblock In \emph{Proceedings of the IEEE/CVF International Conference on
  Computer Vision}, 2023.

\bibitem[Sun et~al.(2021)Sun, Bao, Liu, Fu, Black, and Mei]{sun2021monocular}
Yu Sun, Qian Bao, Wu Liu, Yili Fu, Michael~J Black, and Tao Mei.
\newblock Monocular, one-stage, regression of multiple 3{D} people.
\newblock In \emph{Proceedings of the IEEE/CVF international conference on
  computer vision}, 2021.

\bibitem[Tancik et~al.(2020)Tancik, Srinivasan, Mildenhall, Fridovich-Keil,
  Raghavan, Singhal, Ramamoorthi, Barron, and Ng]{tancik2020fourier}
Matthew Tancik, Pratul Srinivasan, Ben Mildenhall, Sara Fridovich-Keil, Nithin
  Raghavan, Utkarsh Singhal, Ravi Ramamoorthi, Jonathan Barron, and Ren Ng.
\newblock Fourier features let networks learn high frequency functions in low
  dimensional domains.
\newblock \emph{Advances in Neural Information Processing Systems}, 33, 2020.

\bibitem[Thies et~al.(2019)Thies, Zollh{\"o}fer, and
  Nie{\ss}ner]{thies2019deferred}
Justus Thies, Michael Zollh{\"o}fer, and Matthias Nie{\ss}ner.
\newblock Deferred neural rendering: Image synthesis using neural textures.
\newblock \emph{Acm Transactions on Graphics (TOG)}, 38\penalty0 (4):\penalty0
  1--12, 2019.

\bibitem[Tretschk et~al.(2021)Tretschk, Tewari, Golyanik, Zollh{\"o}fer,
  Lassner, and Theobalt]{tretschk2021non}
Edgar Tretschk, Ayush Tewari, Vladislav Golyanik, Michael Zollh{\"o}fer,
  Christoph Lassner, and Christian Theobalt.
\newblock Non-rigid neural radiance fields: Reconstruction and novel view
  synthesis of a dynamic scene from monocular video.
\newblock In \emph{Proceedings of the IEEE/CVF International Conference on
  Computer Vision}, 2021.

\bibitem[Wang et~al.(2022)Wang, Wu, Guo, Zhang, Tai, and Hu]{wang2022NeRF}
Chen Wang, Xian Wu, Yuan-Chen Guo, Song-Hai Zhang, Yu-Wing Tai, and Shi-Min Hu.
\newblock {N}e{R}{F}-sr: High quality neural radiance fields using
  supersampling.
\newblock In \emph{Proceedings of the 30th ACM International Conference on
  Multimedia}, 2022.

\bibitem[Wang et~al.(2021)Wang, Wang, Genova, Srinivasan, Zhou, Barron,
  Martin-Brualla, Snavely, and Funkhouser]{wang2021ibrnet}
Qianqian Wang, Zhicheng Wang, Kyle Genova, Pratul~P Srinivasan, Howard Zhou,
  Jonathan~T Barron, Ricardo Martin-Brualla, Noah Snavely, and Thomas
  Funkhouser.
\newblock Ibrnet: Learning multi-view image-based rendering.
\newblock In \emph{Proceedings of the IEEE/CVF Conference on Computer Vision
  and Pattern Recognition}, 2021.

\bibitem[Weng et~al.(2022)Weng, Curless, Srinivasan, Barron, and
  Kemelmacher-Shlizerman]{weng2022humanNeRF}
Chung-Yi Weng, Brian Curless, Pratul~P Srinivasan, Jonathan~T Barron, and Ira
  Kemelmacher-Shlizerman.
\newblock Human{N}e{R}{F}: Free-viewpoint rendering of moving people from
  monocular video.
\newblock In \emph{Proceedings of the IEEE/CVF Conference on Computer Vision
  and Pattern Recognition}, 2022.

\bibitem[Wu et~al.(2020)Wu, Wang, Hu, and Yu]{wu2020multi}
Minye Wu, Yuehao Wang, Qiang Hu, and Jingyi Yu.
\newblock Multi-view neural human rendering.
\newblock In \emph{Proceedings of the IEEE/CVF Conference on Computer Vision
  and Pattern Recognition}, 2020.

\bibitem[Wu et~al.(2022)Wu, Li, Peng, Lu, Cao, and Zhong]{wu2022dof}
Zijin Wu, Xingyi Li, Juewen Peng, Hao Lu, Zhiguo Cao, and Weicai Zhong.
\newblock Dof-{N}e{R}{F}: Depth-of-field meets neural radiance fields.
\newblock In \emph{Proceedings of the 30th ACM International Conference on
  Multimedia}, 2022.

\bibitem[Xian et~al.(2021)Xian, Huang, Kopf, and Kim]{xian2021space}
Wenqi Xian, Jia-Bin Huang, Johannes Kopf, and Changil Kim.
\newblock Space-time neural irradiance fields for free-viewpoint video.
\newblock In \emph{Proceedings of the IEEE/CVF Conference on Computer Vision
  and Pattern Recognition}, 2021.

\bibitem[Xiu et~al.(2022)Xiu, Yang, Tzionas, and Black]{xiu2022icon}
Yuliang Xiu, Jinlong Yang, Dimitrios Tzionas, and Michael~J Black.
\newblock Icon: Implicit clothed humans obtained from normals.
\newblock In \emph{Proceedings of the IEEE/CVF Conference on Computer Vision
  and Pattern Recognition}, 2022.

\bibitem[Xu et~al.(2018)Xu, Chatterjee, Zollh{\"o}fer, Rhodin, Mehta, Seidel,
  and Theobalt]{xu2018monoperfcap}
Weipeng Xu, Avishek Chatterjee, Michael Zollh{\"o}fer, Helge Rhodin, Dushyant
  Mehta, Hans-Peter Seidel, and Christian Theobalt.
\newblock Monoperfcap: Human performance capture from monocular video.
\newblock \emph{ACM Transactions on Graphics (ToG)}, 37\penalty0 (2):\penalty0
  1--15, 2018.

\bibitem[Yu et~al.(2023)Yu, Cheng, Liu, Wu, and Lin]{yu2023monohuman}
Zhengming Yu, Wei Cheng, Xian Liu, Wayne Wu, and Kwan-Yee Lin.
\newblock Monohuman: Animatable human neural field from monocular video.
\newblock In \emph{Proceedings of the IEEE/CVF Conference on Computer Vision
  and Pattern Recognition}, 2023.

\bibitem[Zhang et~al.(2020)Zhang, Riegler, Snavely, and
  Koltun]{zhang2020NeRF++}
Kai Zhang, Gernot Riegler, Noah Snavely, and Vladlen Koltun.
\newblock {N}e{R}{F}++: Analyzing and improving neural radiance fields.
\newblock \emph{arXiv preprint arXiv:2010.07492}, 2020.

\bibitem[Zhang et~al.(2018)Zhang, Isola, Efros, Shechtman, and
  Wang]{zhang2018unreasonable}
Richard Zhang, Phillip Isola, Alexei~A Efros, Eli Shechtman, and Oliver Wang.
\newblock The unreasonable effectiveness of deep features as a perceptual
  metric.
\newblock In \emph{Proceedings of the IEEE conference on computer vision and
  pattern recognition}, 2018.

\bibitem[Zhang et~al.(2021)Zhang, Srinivasan, Deng, Debevec, Freeman, and
  Barron]{zhang2021NeRFactor}
Xiuming Zhang, Pratul~P Srinivasan, Boyang Deng, Paul Debevec, William~T
  Freeman, and Jonathan~T Barron.
\newblock {N}e{R}{F}actor: Neural factorization of shape and reflectance under
  an unknown illumination.
\newblock \emph{ACM Transactions on Graphics (TOG)}, 40\penalty0 (6):\penalty0
  1--18, 2021.

\bibitem[Zhao et~al.(2022)Zhao, Yang, Zhang, Lin, Zhang, Yu, and
  Xu]{zhao2022humanNeRF}
Fuqiang Zhao, Wei Yang, Jiakai Zhang, Pei Lin, Yingliang Zhang, Jingyi Yu, and
  Lan Xu.
\newblock Human{N}e{R}{F}: Efficiently generated human radiance field from
  sparse inputs.
\newblock In \emph{Proceedings of the IEEE/CVF Conference on Computer Vision
  and Pattern Recognition}, 2022.

\bibitem[Zheng et~al.(2022)Zheng, Huang, Yu, Zhang, Guo, and
  Liu]{zheng2022structured}
Zerong Zheng, Han Huang, Tao Yu, Hongwen Zhang, Yandong Guo, and Yebin Liu.
\newblock Structured local radiance fields for human avatar modeling.
\newblock In \emph{Proceedings of the IEEE/CVF Conference on Computer Vision
  and Pattern Recognition}, 2022.

\bibitem[Zheng et~al.(2023)Zheng, Zhao, Zhang, Liu, and
  Liu]{zheng2023avatarrex}
Zerong Zheng, Xiaochen Zhao, Hongwen Zhang, Boning Liu, and Yebin Liu.
\newblock Avatarrex: Real-time expressive full-body avatars.
\newblock \emph{arXiv preprint arXiv:2305.04789}, 2023.

\bibitem[Zhou et~al.(2021)Zhou, Jiang, Cai, and Lei]{zhou2021dc}
Shihao Zhou, Mengxi Jiang, Shanshan Cai, and Yunqi Lei.
\newblock Dc-gnet: Deep mesh relation capturing graph convolution network for
  3{D} human shape reconstruction.
\newblock In \emph{Proceedings of the 29th ACM International Conference on
  Multimedia}, 2021.

\bibitem[Zhou et~al.(2018)Zhou, Tucker, Flynn, Fyffe, and
  Snavely]{zhou2018stereo}
Tinghui Zhou, Richard Tucker, John Flynn, Graham Fyffe, and Noah Snavely.
\newblock Stereo magnification: Learning view synthesis using multiplane
  images.
\newblock \emph{arXiv preprint arXiv:1805.09817}, 2018.

\end{thebibliography}
}

{
\clearpage
\setcounter{page}{1}
\maketitlesupplementary

\section{Network Architexture}

\subsection{Conv-Filter}
We performed channel-by-channel single-layer convolution on the voxel volume we constructed, with convolution kernel weights initialized to one. Our convolution kernel size is 5, padding is 2, and stride is 1 to keep the volume size.
Channel-by-channel convolution can preserve the semantic information of high-frequency details.
\subsection{Point Refiene}
\begin{figure}[h]
    \centering
    \includegraphics[width=0.33\textwidth]{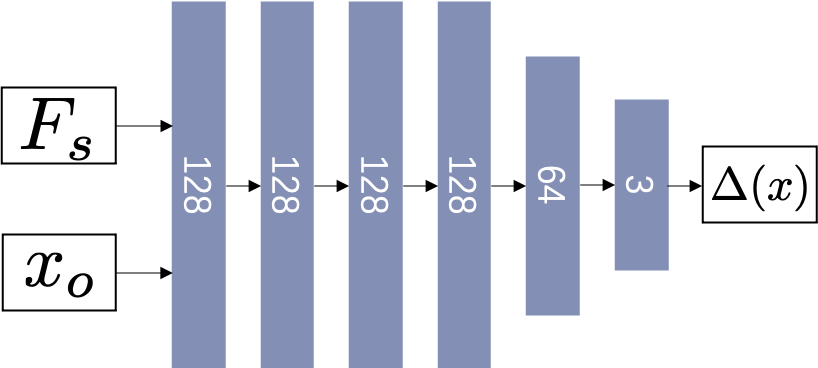}
    \caption{\textbf{Visualization of point refine network.} We use $F_s$ and $x_o$ as inputs to the network to obtain offsets $\Delta(x)$ to refine the canonical coordinates after the general rigid deformation. We initialize the bias of the last layer to zero, and the weight is within the range of $(-1e^{-5},-1e^{-5})$.}
    \label{fig:PointRefiner}
    \vspace{-6mm}
\end{figure}

\subsection{NeRF Network}
\begin{figure}[h]
    \centering
    \includegraphics[width=0.4\textwidth]{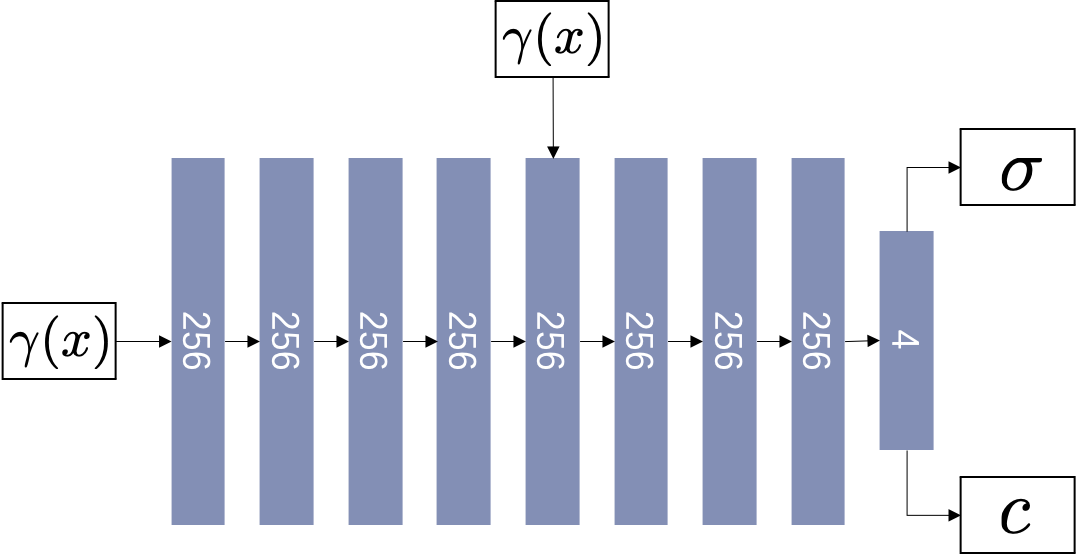}
    \caption{\textbf{Visualization of appearance network.} Follow the baseline~\cite{weng2022humanNeRF}, we use an 8-layer MLP with width=256, taking as input positional encoding $\gamma$ of position $x$ and producing color $c$ and density $\sigma$. A skip connection that concatenates $\gamma(x)$ to the fifth layer is applied. We adopt ReLU activation after each fully connected layer, except for the one generating color $c$ where we use sigmoid.}
    \label{fig:NeRFnetwork}
    \vspace{-6mm}
\end{figure}

\section{Canonicalization}
To map points in the observation space to the canonical space, our baselines utilize neural networks to learn backward warping weight fields. This method is easy to implement but suffers from poor generalization to unseen poses, as the backward weight fields attempt to learn a spatial weight fields that deform with pose variations, necessitating memorization of weight fields for different spatial configurations. Generalizing to unseen poses using such pose-dependent weight fields is difficult. 
Our method directly queries the nearest SMPL's LBS weights in an explicit voxel, which is highly efficient and generalizable. However, this method suffers from unnatural deformations when the deformation angle is too large (see Figure~\ref{fig:pointrefiner}). To address this issue, we use an additional neural network to learn a residual for canonical points, which is specific to the data and empowered by point-level features, considering the different clothing of the performers.

The rotation $R_j$ and translation $T_j$ for the rigid deformation are represented as:
\begin{equation}
\begin{bmatrix}
R_j & T_j \\ 
 0&1 
\end{bmatrix}
=
\prod_{i \in p(j)}
\begin{bmatrix}
R(\omega_i^c ) &o_i^c \\ 
 0&1 
\end{bmatrix}
\begin{Bmatrix}
\prod_{i \in p(j)}
\begin{bmatrix}
R(\omega_i ) &o_i \\ 
 0&1 
\end{bmatrix}
\end{Bmatrix}^{-1}
\end{equation}
where $p(j)$ is the ordered set of parents of joint $j$ in the kinematic tree, $\omega_i$ defines local joint rotations using axis-angle representations, $R(\omega_i) \in \mathbb{R}^{3\times 3}$ is the converted rotation matrix of $\omega_i$ via the Rodrigues formula, and $o_i$ is the $i$-th joint center. 

\section{Ablation Study}
\paragraph{Voxel Size.}
In the Conv-Filter, we voxelized all coordinates, which resulted in some loss of information compared to a dense space. However, this greatly facilitated our subsequent processing and computations. We conducted ablation experiments on the voxel size used in the voxelization process, as shown in Table~\ref{table:ablation-voxelsize}. 
Voxel size affects mapping granularity to get canonical points. 
A larger voxel size results in coarser mapping and lower image quality. 
In contrast, a smaller one requires a bigger convolution kernel to diffuse occupancy and more computing resources, and it did not result in higher accuracy because the prior does not perfectly match the actual human body, and clothing is also outside the prior.

\begin{table}[t]
\centering
\footnotesize
\caption{Ablation experiment on the voxel size of our method in \textsc{zju-mocap}. For different subjects, there is a different optimal voxel size. We use 0.02 as voxel size in the experiment section because it has the best overall performance.}
\begin{tabular}{llll}
\toprule
 & \multicolumn{1}{c}{\small{PSNR$\uparrow$}} & \multicolumn{1}{c}{\small{SSIM$\uparrow$}} & \multicolumn{1}{c}{$^\dagger$LPIPS$\downarrow$} \\ 
\midrule
 voxel=0.01   & 30.49 & 0.971 & 26.386 \\
 voxel=0.015   & 30.69 & 0.972 & 25.354 \\
 voxel=0.02  & 31.09 \gold & 0.974 \gold & 24.085 \gold \\
 voxel=0.025   & 31.07 & 0.974 \gold & 24.705 \\
 voxel=0.03   & 30.72 & 0.973 & 25.013 \\
 \bottomrule
\end{tabular}
\vspace{-6mm}
\label{table:ablation-voxelsize}

\end{table}

\paragraph{Weight Distribution.}
In the main text, we discussed the importance of filtering operations. It is worth noting that previous methods did not require similar operations. Our experiments suggest that this is because the learned neural weight field assigns negative weights to irrelevant joints in the data. When we simply use $softmax$ to map the neural weight field to a distribution that is the same as the SMPL weight, $\sum_{j \in J}w_j = 1, w_j>0$, similar phenomena occur, see Figure~\ref{fig:su-weightdistribution}.

\begin{figure}
    \centering
    \includegraphics[width=0.4\textwidth]{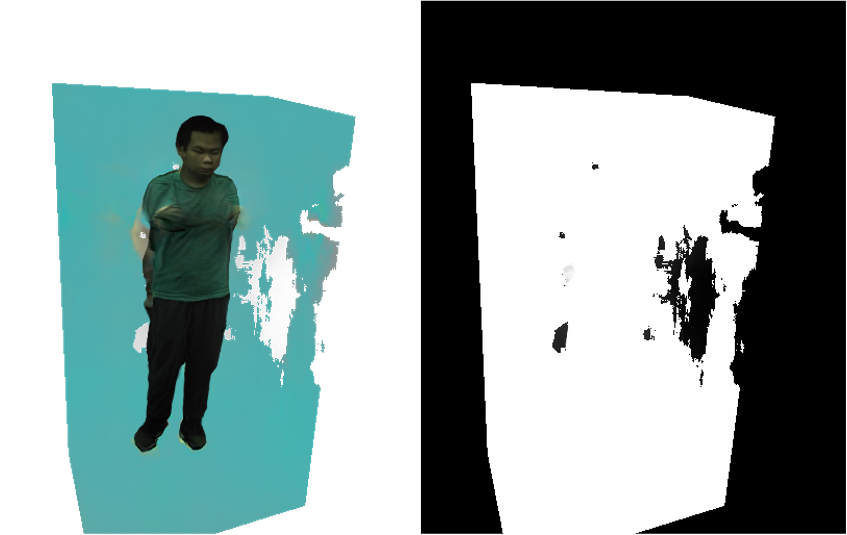}
    \caption{By mapping the learned weight field of HumanNeRF~\cite{weng2022humanNeRF} to the same distribution as SMPL weight using softmax, the same phenomenon occurred.}
    \label{fig:su-weightdistribution}
    \vspace{-4mm}
\end{figure}

\section{More Results}
\paragraph{Quantitative experiments.} Due to the space limitation of the main text, the Figure~\ref{fig:quantitative} provided in the main text is the few-shot image synthesis result and Table~\ref{tab:quantitive-results} is a overall table.

Table~\ref{tab:su-fullinput-zjumocap}, Table~\ref{tab:su-fewshot-zjumocap} and Table~\ref{tab:su-custom} are detailed data for Table~\ref{tab:quantitive-results}. For \textsc{zju-mocap}, we use videos from the same six subjects as in previous work. For our \textsc{in-the-wild data}, each subject is composed of multiple pose sequences to enable more convincing novel pose experiments. The videos are captured with a single camera, and SMPL are estimated with ROMP~\cite{sun2021monocular}.

In order to better compare the performance of different methods, we synthesized the results using full input images. Figure~\ref{fig:su-fullzjumocap} shows the synthesis results on \textsc{ZJU-MoCap}, and Figure~\ref{fig:su-fullcustom} shows the image synthesis results on the \textsc{in-the-wild data}. Figure~\ref{fig:su-full-fewshot} directly compares the novel pose image synthesis ability of the three methods under two input conditions.

Compared with Figure~\ref{fig:quantitative}, the defects of the baselines have been significantly improved on \textsc{ZJU-MoCap}, but there are still obvious defects in these methods on \textsc{in-the-wild data}. This is highly related to the data distribution.

\vspace{-3mm}
\paragraph{Pose similarity.} We found that the novel pose experimental setting in previous studies is unreasonable because the test poses and training poses are very similar, as the characters in \textsc{zju-mocap} perform similar movements repeatedly. This also encourages us to use custom data and subjective research of results on poses without ground truth images. Novel poses synthesized by HumanNeRF should be sufficiently novel and easy to obtain, rather than being limited to professional laboratories.

To quantify the similarity of poses,
we calculate the highest cosine similarity with all training poses for each pose in the test data. As shown in Table~\ref{tab:posesimi}, simply dividing each subject into training and testing data in a 4:1 ratio may result in highly similar poses in the testing data being present in the training data.

In our experiments, high pose similarity does not always result in a decrease in baseline performance, because during the training process, similar poses and viewpoints are limited. For example, the pose similarity of Subject 386 is very high, but the corresponding pose only appears at the beginning and can only see the performer's right side, so when we synthesize this highly similar pose, the performance of the baselines is not good.
\begin{table}[!t]
\centering
\footnotesize
\caption{Pose similarity of test poses and train poses in previous novel pose experiments. The previous setting of novel pose is not novel enough because of the high similarity. }
\begin{tabular}{cccc}
    & Min   & Max   & Average     \\ \hline
377 & 0.876 & 0.997 & 0.919       \\
386 & 0.939 & 0.995 & 0.964       \\
387 & 0.969 & 0.996 & 0.985       \\
392 & 0.845 & 0.957 & 0.909       \\
393 & 0.834 & 0.998 & 0.904       \\
394 & 0.760 & 0.993 & 0.842      
\end{tabular}
\label{tab:posesimi}
\vspace{-7mm}
\end{table}

\vspace{-4mm}
\paragraph{Qualitative results.} In the qualitative experiments, we drive the model to generate new pose images by using pose sequences of different performers. We make a series of image results into a video to help the human eyes better distinguish the performance of different methods.
The video results are included in the supplementary materials in the form of a compressed file.

\vspace{-4mm}
\paragraph{Visual quality.} The rendering results and metrics on the \textsc{ZJU-Mocap} data are 512p resolution which we followed the popular protocol in \cite{yu2023monohuman,weng2022humanNeRF}, while on our collected \textsc{in-the-wild} data the resolution is 1080p. Both results show our superiority over SOTAs. To validate the reliability of these scores calculated with 512p, we compare our method with HumanNeRF on frames with 2K resolution and find that we still surpass HumanNeRF, as shown in Table~\ref{tab:highres}.
We provide additional visual comparisons based on higher-resolution frames (Figure~\ref{fig:highres}). The logo and hand show better details than HumanNeRF in both low and high resolution.

\begin{table}[t]
     \centering
     \caption{Results on subject 392 under different resolution.}
     \label{tab:highres}
\resizebox{\columnwidth}{!}
{
\begin{tabular}{lccclccc}
\toprule
\multicolumn{1}{c}{2k resolution} & \multicolumn{1}{c}{{PSNR$\uparrow$}} & \multicolumn{1}{c}{{SSIM$\uparrow$}} & \multicolumn{1}{c}{LPIPS$\downarrow$}
&\multicolumn{1}{c}{ 512p resolution} & \multicolumn{1}{c}{{PSNR$\uparrow$}} & \multicolumn{1}{c}{{SSIM$\uparrow$}} & \multicolumn{1}{c}{LPIPS$\downarrow$}\\ 
\cmidrule(lr){1-4}\cmidrule(lr){5-8}
HumanNeRF  & 31.36  & 0.983 & 0.0318 &HumanNeRF &\textbf{31.55} &\textbf{0.975} & 0.0280\\
Ours & \textbf{31.73}  & \textbf{0.984}  & \textbf{0.0314} &Ours &31.36 &0.973 &\textbf{0.0276} \\
\bottomrule
\end{tabular}}
\vspace{-3mm}
\end{table}

\vspace{-6pt}
\begin{figure}[h]
    \centering
    \includegraphics[width=1\linewidth]{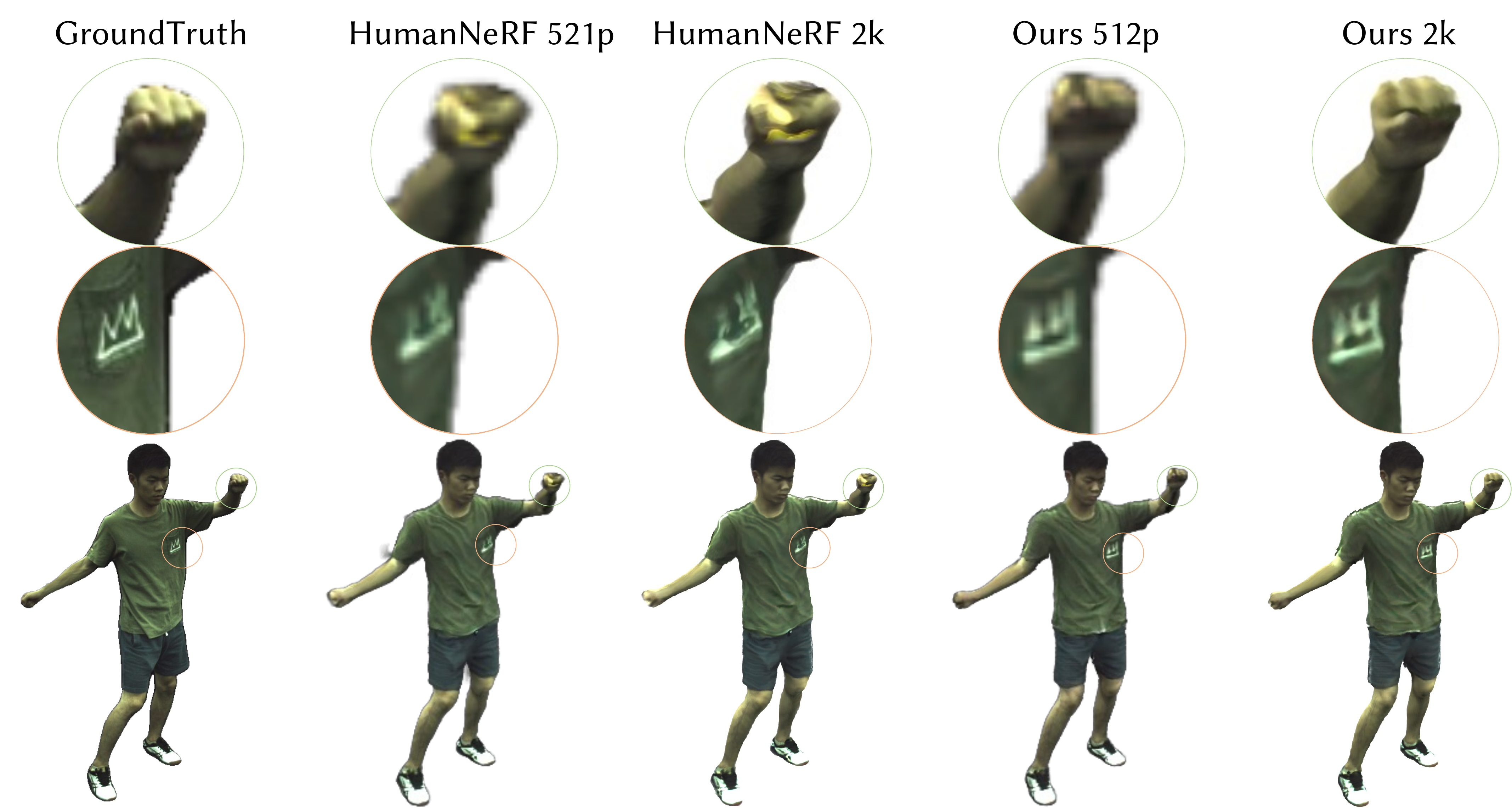}
    \vspace{-18pt}
    \caption{Results on subject 392 under different resolutions.}
    \label{fig:highres}
\end{figure}


\paragraph{More comparisons.} To demonstrate the excellent performance of our method in similar tasks, additional comparisons with methods that incorporate explicit information are provided in Table~\ref{table:Thuman}.

\begin{table}[t]
\centering
\caption{ Comparison with more methods. Our method has shown significant advantages in comparison with more methods.}
\vspace{-1mm}
\resizebox{\columnwidth}{!}
{
 \renewcommand\tabcolsep{10pt}
\begin{tabular}{clccccccc}
\toprule
  & \multicolumn{1}{c}{} & View & Prior & \multicolumn{1}{c}{{PSNR$\uparrow$}} & \multicolumn{1}{c}{{SSIM$\uparrow$}} & \multicolumn{1}{c}{LPIPS$\downarrow$} \\ 
\midrule[0.3pt]
\multirow{5}{*}{\rotatebox{90}{\textsc{Thuman4.0}}} & TAVA~\cite{li2022tava} & $>$1 & skeleton & 26.607  & 0.968  & 0.032 \\
& SLRF~\cite{zheng2022structured} & 24 & nodes & 26.152 & 0.969 & 0.024 \\ 
&  Posevocab~\cite{li2023posevocab} & 24 & SMPL & 30.972   &0.977   &\textbf{0.017}  \\
\cmidrule(lr){2-7}
& Posevocab~\cite{li2023posevocab} &1 & SMPL &27.820   &0.973   &0.064  \\
& Ours &1 &SMPL & \textbf{31.148}  & \textbf{0.979}  & \textbf{0.017} \\
\midrule
\midrule
\multirow{4}{*}{\rotatebox{90}{\textsc{ZJU-Mocap}}} & SLRF~\cite{zheng2022structured}  & 24 & nodes & 23.61 & 0.905 & -- \\ 
& NPC~\cite{su2023npc} &$>$1 &point clouds & 21.88   & --   & 0.134  \\\cmidrule(lr){2-7}
& SelfRecon~\cite{jiang2022selfrecon} &1 & SMPL & 27.94  & 0.969 &0.043 \\\cmidrule(lr){2-7}
& Ours &1 &SMPL & \textbf{29.36}  & \textbf{0.974}  & \textbf{0.022} \\
\bottomrule
\end{tabular}
}
\label{table:Thuman}

\end{table}

\paragraph{Dancing visualization.} In addition to using cross-subject movements in our qualitative analysis to test the model's novel pose ability, we can also use dance movements from online videos to drive the model as presented in Figure~\ref{fig:dacing}.
\vspace{-6pt}
\begin{figure}[h]
    \centering
    \includegraphics[width=1\linewidth]{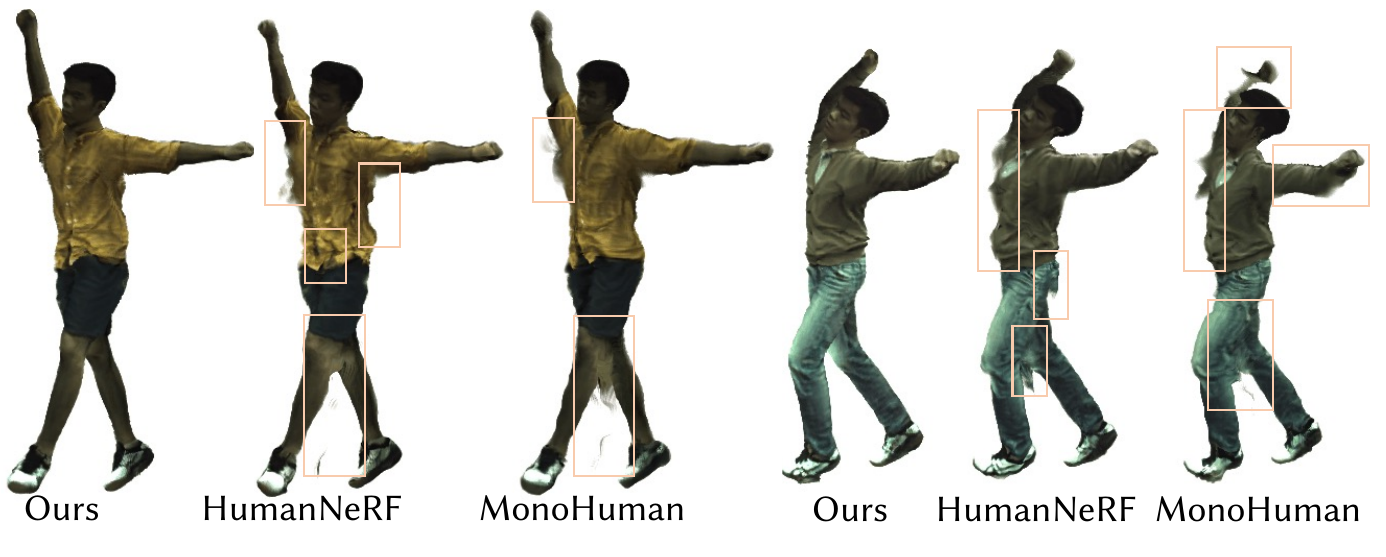}
    \vspace{-18pt}
    \caption{Dancing results. Our results are very clean compared to the blurry and unnatural results of other methods.}
    \label{fig:dacing}
    \vspace{-5mm}
\end{figure}
\begin{table*}[!h]
\centering
\caption{\textbf{Comparison of full input results on \textsc{zju-mocap}.} Our method shows a leading advantage in LPIPS, which is aligned with human perception. Each data in the table represents the average value of all test frames' metrics in the corresponding video, which is consistent with the previous studies.}
\footnotesize

\begin{minipage}{\textwidth}
\centering
\begin{tabular}{llllllllll}
\toprule
\multicolumn{1}{c}{\multirow{2}{*}{Methods}} & \multicolumn{3}{c}{Subject377}    & \multicolumn{3}{c}{Subject386}    & \multicolumn{3}{c}{Subject387}\\ 
\cmidrule(lr){2-4}\cmidrule(lr){5-7}\cmidrule(lr){8-10}
& \multicolumn{1}{c}{\small{PSNR$\uparrow$}} & \multicolumn{1}{c}{\small{SSIM$\uparrow$}} & \multicolumn{1}{c}{\small{$^\dagger$LPIPS$\downarrow$}} & \multicolumn{1}{c}{\small{PSNR$\uparrow$}} & \multicolumn{1}{c}{\small{SSIM$\uparrow$}} & \multicolumn{1}{c}{\small{$^\dagger$LPIPS$\downarrow$}} & \multicolumn{1}{c}{\small{PSNR$\uparrow$}} & \multicolumn{1}{c}{\small{SSIM$\uparrow$}} & \multicolumn{1}{c}{\small{$^\dagger$LPIPS$\downarrow$}}\\
\midrule
Ani-NeRF~\cite{peng2021animatable} &22.12 &0.8790 &51.796 &23.92 &0.8545 &55.697 &16.26 &0.7850 &89.790 \\
HumanNeRF~\cite{weng2022humanNeRF} & {30.74}  & {0.9795}  & 17.387 & {33.46}  & {0.9716}  & {36.326} & 30.30 & 0.9768 \gold & 20.010 \gold \\
MonoHuman~\cite{yu2023monohuman} & 31.82 \gold & 0.9822 & 17.561 & 30.10 & 0.9561 & 69.107 & {30.43} \gold  & {0.9755}  & 23.954  \\
\rowcolor{CBLightGreen} Ours  & 31.34 & 0.9826 \gold & {15.129} \gold & 33.80 \gold & 0.9741 \gold & 32.990 \gold  & 29.36 & 0.9742 & {21.462} 
\end{tabular}
\end{minipage}
\rule{0pt}{1ex}
\begin{minipage}{\textwidth}
\centering
\begin{tabular}{llllllllll}
\toprule
\multicolumn{1}{c}{\multirow{2}{*}{Methods}} & \multicolumn{3}{c}{Subject392}    & \multicolumn{3}{c}{Subject393}    & \multicolumn{3}{c}{Subject394}  \\ 
\cmidrule(lr){2-4}\cmidrule(lr){5-7}\cmidrule(lr){8-10}
& \multicolumn{1}{c}{\small{PSNR$\uparrow$}} & \multicolumn{1}{c}{\small{SSIM$\uparrow$}} & \multicolumn{1}{c}{\small{$^\dagger$LPIPS$\downarrow$}} & \multicolumn{1}{c}{\small{PSNR$\uparrow$}} & \multicolumn{1}{c}{\small{SSIM$\uparrow$}} & \multicolumn{1}{c}{\small{$^\dagger$LPIPS$\downarrow$}} & \multicolumn{1}{c}{\small{PSNR$\uparrow$}} & \multicolumn{1}{c}{\small{SSIM$\uparrow$}} & \multicolumn{1}{c}{\small{$^\dagger$LPIPS$\downarrow$}}\\
\midrule
Ani-NeRF~\cite{peng2021animatable} &22.78 &0.8610 &67.744 &20.37 &0.8417 & 75.381 &22.05 &0.8537 &69.188 \\
HumanNeRF~\cite{weng2022humanNeRF} & {31.80}  & {0.9743} & 26.811 & {29.80} \gold & {0.9708} \gold & {25.615} & 30.85 & 0.9702 & 22.783 \gold  \\
MonoHuman~\cite{yu2023monohuman} & 32.06 \gold & 0.9749 \gold & 27.043 & 29.69 & 0.9701 & 26.570 & {31.37} \gold & {0.9720} \gold & 23.521 \\
\rowcolor{CBLightGreen} Ours  & 31.75 & 0.9740 & {25.537} \gold & 29.50 & 0.9642 & 25.462 \gold & 30.81 & 0.9697 & {23.929} \\
\bottomrule
\end{tabular}
\end{minipage}
\label{tab:su-fullinput-zjumocap}
\end{table*}

\begin{table*}[!h]
\centering
\caption{\textbf{Comparison of few-shot input results on \textsc{zju-mocap}.} On the most important metric LPIPS, our method demonstrates the best results. our method exhibits less performance degradation with few-shot input indicates that our method does not overly rely on data to fit the model, unlike previous methods.}
\footnotesize

\begin{minipage}{\textwidth}
\centering
\begin{tabular}{llllllllll}
\toprule
\multicolumn{1}{c}{\multirow{2}{*}{Methods}} & \multicolumn{3}{c}{Subject377}    & \multicolumn{3}{c}{Subject386}    & \multicolumn{3}{c}{Subject387}\\ 
\cmidrule(lr){2-4}\cmidrule(lr){5-7}\cmidrule(lr){8-10}
& \multicolumn{1}{c}{\small{PSNR$\uparrow$}} & \multicolumn{1}{c}{\small{SSIM$\uparrow$}} & \multicolumn{1}{c}{\small{$^\dagger$LPIPS$\downarrow$}} & \multicolumn{1}{c}{\small{PSNR$\uparrow$}} & \multicolumn{1}{c}{\small{SSIM$\uparrow$}} & \multicolumn{1}{c}{\small{$^\dagger$LPIPS$\downarrow$}} & \multicolumn{1}{c}{\small{PSNR$\uparrow$}} & \multicolumn{1}{c}{\small{SSIM$\uparrow$}} & \multicolumn{1}{c}{\small{$^\dagger$LPIPS$\downarrow$}}\\
\midrule
Ani-NeRF~\cite{peng2021animatable} &21.77 &0.8313 &66.532 &24.48 &0.8386 &74.342 &20.74 &0.8279 &65.747 \\
HumanNeRF~\cite{weng2022humanNeRF} & 30.33    & 0.9799   & 18.510   & 31.47    & 0.9618   & 48.410   & 27.92    & 0.9610    & 39.941 \\
MonoHuman~\cite{yu2023monohuman} & 31.02    & 0.9774   & 22.562   & 31.26    & 0.9601   & 56.916   & 28.5 \gold    & 0.9619 \gold  & 43.147  \\
\rowcolor{CBLightGreen} Ours  & 31.07 \gold   & 0.9806 \gold  & 18.484 \gold  & 32.44 \gold   & 0.9644 \gold  & 43.876 \gold  & 28.05    & 0.9612   & 39.733  \gold
\end{tabular}
\end{minipage}
\rule{0pt}{1ex}
\begin{minipage}{\textwidth}
\centering
\begin{tabular}{llllllllll}
\toprule
\multicolumn{1}{c}{\multirow{2}{*}{Methods}} & \multicolumn{3}{c}{Subject392}    & \multicolumn{3}{c}{Subject393}    & \multicolumn{3}{c}{Subject394}  \\ 
\cmidrule(lr){2-4}\cmidrule(lr){5-7}\cmidrule(lr){8-10}
& \multicolumn{1}{c}{\small{PSNR$\uparrow$}} & \multicolumn{1}{c}{\small{SSIM$\uparrow$}} & \multicolumn{1}{c}{\small{$^\dagger$LPIPS$\downarrow$}} & \multicolumn{1}{c}{\small{PSNR$\uparrow$}} & \multicolumn{1}{c}{\small{SSIM$\uparrow$}} & \multicolumn{1}{c}{\small{$^\dagger$LPIPS$\downarrow$}} & \multicolumn{1}{c}{\small{PSNR$\uparrow$}} & \multicolumn{1}{c}{\small{SSIM$\uparrow$}} & \multicolumn{1}{c}{\small{$^\dagger$LPIPS$\downarrow$}}\\
\midrule
Ani-NeRF~\cite{peng2021animatable} &22.49 &0.8433 &61.917 &21.96 &0.8384 & 59.169 &21.65 &0.8240 &61.328 \\
HumanNeRF~\cite{weng2022humanNeRF} & 31.55 \gold   & 0.9747 \gold  & 28.043   & 29.45 \gold   & 0.9694 \gold  & 26.930 \gold  & 28.66    & 0.9629   & 36.507   \\
MonoHuman~\cite{yu2023monohuman} & 31.48    & 0.9739   & 30.090   & 29.45 \gold   & 0.9691   & 30.113   & 28.86 \gold   & 0.9636 \gold  & 36.139 \\
\rowcolor{CBLightGreen} Ours  & 31.36    & 0.9734   & 27.604 \gold  & 29.25    & 0.9681   & 27.328   & 28.51    & 0.9626   & 35.479 \gold \\
\bottomrule
\end{tabular}
\end{minipage}
\label{tab:su-fewshot-zjumocap}
\end{table*}

\begin{table*}[!h]
\centering
\caption{\textbf{Comparison of results on the \textsc{in-the-wild data}.} Since \textsc{in-the-wild data} is not captured in laboratory conditions and only contains monocular information, the accuracy of SMPL estimation is lower compared to \textsc{zju-mocap}, which leads to a decrease in overall performance metrics. However, this is more in line with real-world application scenarios. Our method demonstrates the best performance, especially in the LPIPS metric, which reflects image quality the most. Even with reduced input data, our method maintains excellent performance, while other methods experience a greater degree of decline.}
\footnotesize
\begin{minipage}{\textwidth}
\centering
\begin{tabular}{llllllllll}
\toprule
\multicolumn{1}{c}{\multirow{2}{*}{Full Input}} 
& \multicolumn{3}{c}{S1}    & \multicolumn{3}{c}{S2}    & \multicolumn{3}{c}{S3}  \\ 
\cmidrule(lr){2-4}\cmidrule(lr){5-7}\cmidrule(lr){8-10}
& \multicolumn{1}{c}{\small{PSNR$\uparrow$}} & \multicolumn{1}{c}{\small{SSIM$\uparrow$}} & \multicolumn{1}{c}{\small{$^\dagger$LPIPS$\downarrow$}} & \multicolumn{1}{c}{\small{PSNR$\uparrow$}} & \multicolumn{1}{c}{\small{SSIM$\uparrow$}} & \multicolumn{1}{c}{\small{$^\dagger$LPIPS$\downarrow$}} & \multicolumn{1}{c}{\small{PSNR$\uparrow$}} & \multicolumn{1}{c}{\small{SSIM$\uparrow$}} & \multicolumn{1}{c}{\small{$^\dagger$LPIPS$\downarrow$}}\\
\midrule
HumanNeRF~\cite{weng2022humanNeRF} & 32.94 & 0.9737 & 43.259 & 26.65 & 0.9650 & 41.505 & 27.32 & 0.9500 & 59.620 \\
MonoHuman~\cite{yu2023monohuman} & 32.99 & 0.9725 & 46.272 & 26.74 & 0.9683 & 42.198 & 27.72 \gold & 0.9509 & 66.400  \\
\rowcolor{CBLightGreen} Ours  & 33.72 \gold & 0.9764 \gold & 42.343 \gold & 26.43 \gold & 0.9709 \gold & 39.174 \gold & 27.54 & 0.9524 \gold & 57.408 \gold 
\end{tabular}

\end{minipage}
\rule{0pt}{1ex}
\begin{minipage}{\textwidth}
\centering
\begin{tabular}{llllllllll}
\toprule
\multicolumn{1}{c}{\multirow{2}{*}{Few-shot Input}} 
& \multicolumn{3}{c}{S1}    & \multicolumn{3}{c}{S2}    & \multicolumn{3}{c}{S3}  \\ 
\cmidrule(lr){2-4}\cmidrule(lr){5-7}\cmidrule(lr){8-10}
& \multicolumn{1}{c}{\small{PSNR$\uparrow$}} & \multicolumn{1}{c}{\small{SSIM$\uparrow$}} & \multicolumn{1}{c}{\small{$^\dagger$LPIPS$\downarrow$}} & \multicolumn{1}{c}{\small{PSNR$\uparrow$}} & \multicolumn{1}{c}{\small{SSIM$\uparrow$}} & \multicolumn{1}{c}{\small{$^\dagger$LPIPS$\downarrow$}} & \multicolumn{1}{c}{\small{PSNR$\uparrow$}} & \multicolumn{1}{c}{\small{SSIM$\uparrow$}} & \multicolumn{1}{c}{\small{$^\dagger$LPIPS$\downarrow$}}\\
\midrule
HumanNeRF~\cite{weng2022humanNeRF} &33.37 & 0.9761 & 45.012 & 25.82 & 0.9599 & 43.326 & 27.27 & 0.9495 & 62.384 \\
MonoHuman~\cite{yu2023monohuman}  &33.58 & 0.9744 & 50.561 & 26.57 \gold & 0.9640 & 45.955 & 27.48 & 0.9522 & 72.143 \\
\rowcolor{CBLightGreen} Ours &33.70 \gold & 0.9768 \gold & 41.954 \gold & 26.45 & 0.9712 \gold & 39.894 \gold & 27.62 \gold & 0.9525 \gold & 59.636 \gold   \\
\bottomrule
\end{tabular}
\end{minipage}
\label{tab:su-custom}
\end{table*}

\begin{figure*}
    \vspace{-5pt}
    \centering
    \includegraphics[width=0.9\textwidth]{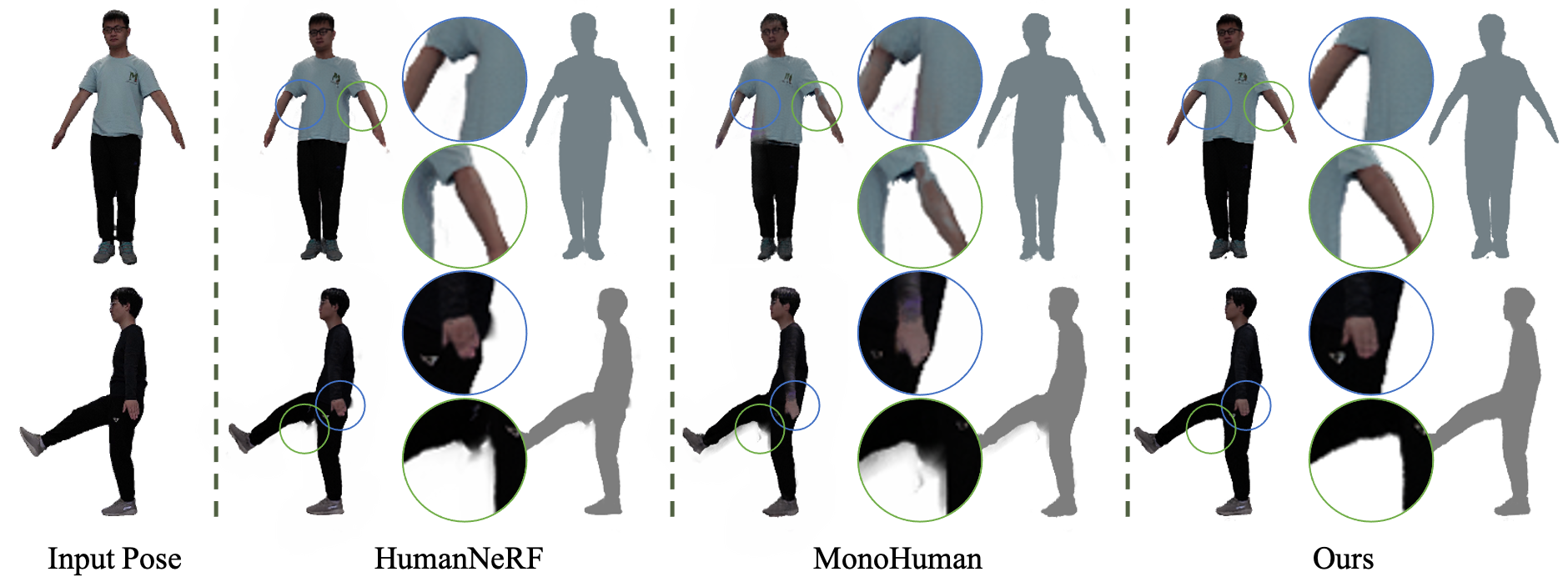}
    \vspace{-10pt}
    \caption{\textbf{Image synthesis results with full input on \textsc{in-the-wild data}.} Our method maintains good performance at joint junctions, but the results of the baselines are blurry and have unnatural distortions.}
    \label{fig:su-fullcustom}
    \vspace{-10pt}
\end{figure*}

\begin{figure*}
    \centering
    \includegraphics[width=0.9\textwidth]{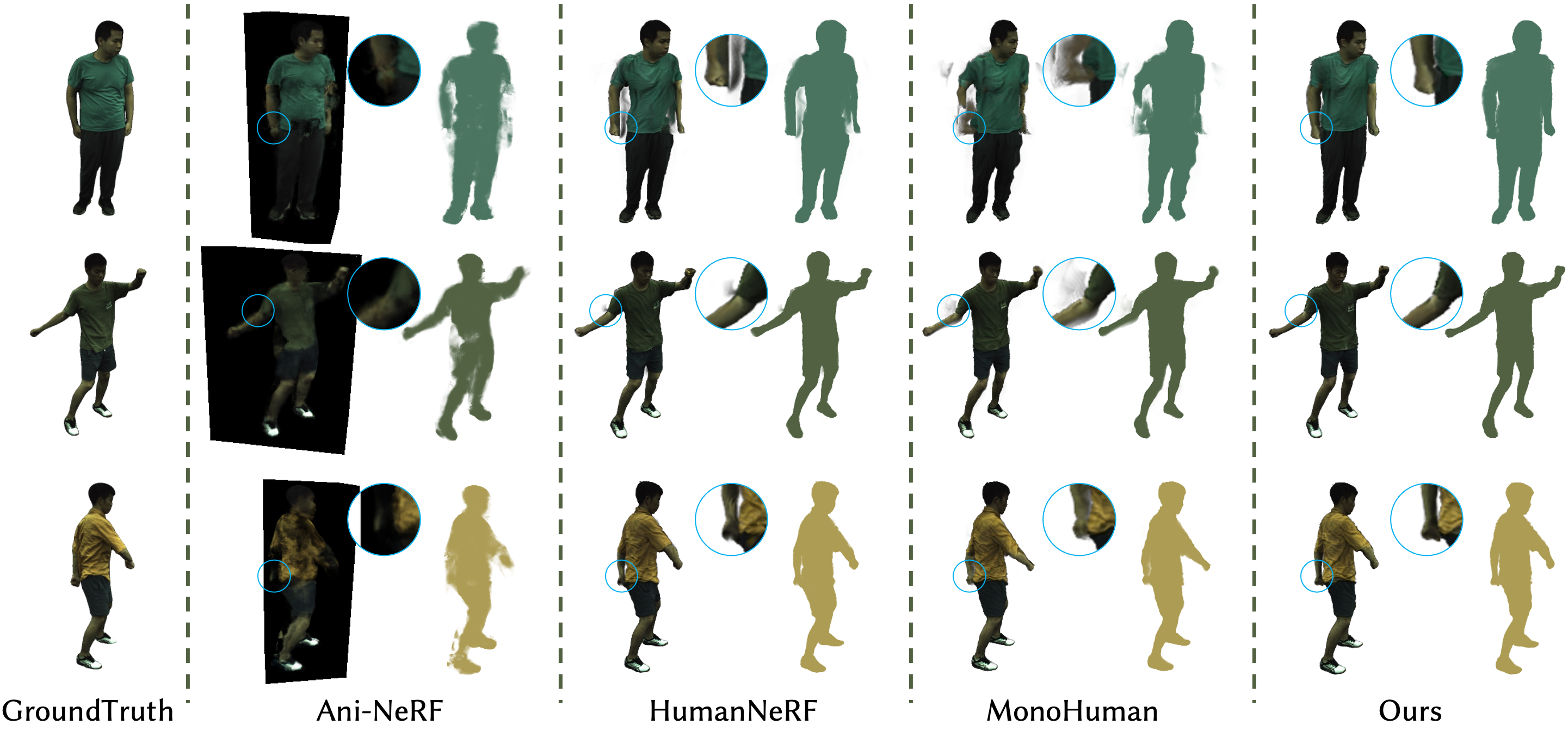}
    \vspace{-10pt}
    \caption{\textbf{Image synthesis results with full input on \textsc{zju-mocap}.} For Subject 386 (line 1), the baselines still have very poor image synthesis results. For Subject 392 and 393, there are still irregular deformations and artifacts in the image synthesis results, whereas our method achieves the best performance in visual comparison.}
    \label{fig:su-fullzjumocap}
\end{figure*}

\begin{figure*}
    \centering
    \includegraphics[width=0.7\textwidth]{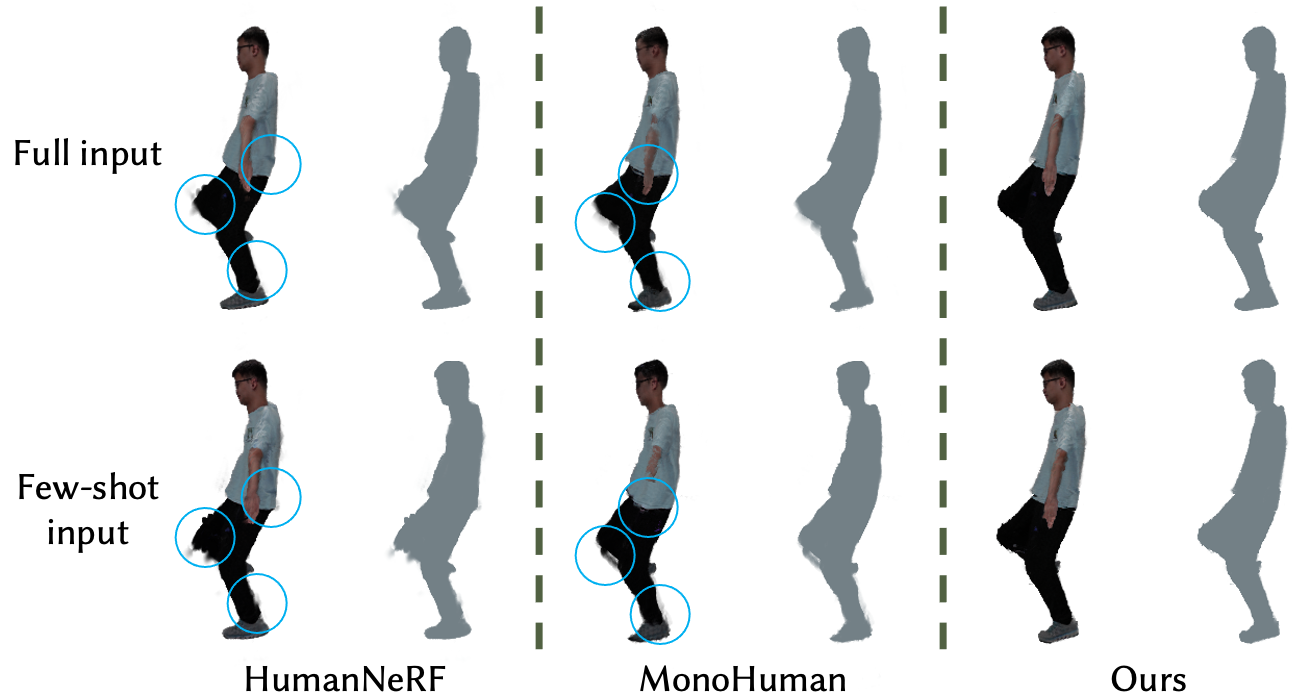}
    \vspace{-10pt}
    \caption{\textbf{Comparison between full input and few-shot input.} HumanNeRF~\cite{weng2022humanNeRF} and MonoHuman~\cite{yu2023monohuman} exhibit more artifacts and unnatural deformations with less data. MonoHuman~\cite{yu2023monohuman} even produces hand missing. Our method maintains almost unchanged performance under the same testing conditions.}
    \label{fig:su-full-fewshot}
    \vspace{-15pt}
    
\end{figure*}
}
\end{document}